\documentclass[11pt]{article}

\usepackage[final]{acl}

\usepackage{times}
\usepackage{latexsym}
\usepackage{tabularx}
\usepackage{booktabs}
\usepackage{multicol, multirow}
\usepackage[table]{xcolor}
\usepackage{array, enumitem}
\usepackage[most, breakable, skins]{tcolorbox}
\usepackage{hyperref}
\usepackage{threeparttable}
\usepackage{siunitx}
\usepackage{makecell}

\usepackage[T1]{fontenc}
\usepackage[utf8]{inputenc}

\usepackage{microtype}

\usepackage{inconsolata}

\usepackage{graphicx}

\title{Sense and Sensitivity: Benchmarking LLM Clinical Triage Recommendations with Physician Experts}

\author{
Abinitha Gourabathina$^{1,*}$ \And
Haoran Zhang$^{1}$ \And
Yuexing Hao$^{1}$
\AND
Walter Gerych$^{2}$ \And
Marzyeh Ghassemi$^{1}$
\AND
\makebox[\textwidth][c]{%
\begin{minipage}{0.95\textwidth}
\centering
\small\mdseries
$^{1}$Massachusetts Institute of Technology, Department of Electrical Engineering and Computer Science \\
$^{2}$Worcester Polytechnic Institute, Department of Computer Science \\
$^{*}$Correspondence: \texttt{abinitha@mit.edu}
\end{minipage}
}
}

\begin{document}
\maketitle
\begin{abstract}
As large language models (LLMs) are increasingly used in clinical settings, it is critical to evaluate their reliability under realistic variation in clinical text. We study this question in clinical triage, comparing LLMs to practicing physicians under text perturbations that preserve the underlying clinical setting. We introduce a benchmark of over $6{,}000$ clinical scenarios, $7{,}000$ physician annotations, and $225{,}000$ model responses. Using this benchmark, we make two key observations. First, LLMs are more likely than physicians to recommend unnecessary care at baseline, and this tendency increases under perturbed inputs. Further, we find that LLM recommendations are more sensitive to gender and tone perturbations than human recommendations. Together, these results demonstrate that LLMs can vary under clinically irrelevant textual changes, highlighting the need for deployment-oriented evaluations grounded in expert physician behavior.
\end{abstract}

\section{Introduction}
Medical Large Language Models (LLMs) have gained widespread attention in recent years for impressive performance in a growing number of clinical tasks, such as summarization of medical information \cite{tang2023evaluating, van2024adapted, de2025we}, disease diagnosis \cite{panagoulias2024evaluating, holmes2025radonc, cai2020dual, cao2023intelligent}, creation of treatment plans for various pathologies \cite{qiu2024llm, li2023meddm}, and encoding clinical knowledge for medical reasoning \cite{singhal2023large, singhal2025toward, hao2025large}. These models have promise in democratizing healthcare \cite{rubeis2022democratizing, turchi2024pathways, sayres2025towards} and reducing physician burden \cite{challenges2023catalyzing, gandhi2023can, olson2025use, hao2025retrospective}. 

Patients are increasingly turning to AI chatbots for clinical recommendations \cite{hao2024advancing, shahsavar2023user, hopkins2023artificial, hao2025personalizing}. Similarly, clinicians are also using chatbots for clinical assistance, ranging from administrative tasks \cite{sorensen2023machine} to  diagnostic assistance \cite{tiwari2025current, cabral2025future}. Thus, it is critical to study how model-generated clinical recommendations respond to realistic textual variation before deployment.

Despite this promise, prior work shows that clinical LLMs can be brittle in ways that differ from human reasoning~\cite{bedi2025fidelity, hao2025medpair, modi2026large, gourabathina}, raising concerns about reliability in real-world settings~\cite{xu2025artificial}. For instance, on medical licensing exam benchmarks~\cite{gilson2023does, pal2024open, li2025medguide}, replacing the correct answer with ``None of the other answers'' substantially reduces clinical LLM accuracy~\cite{bedi2025fidelity}. Real clinical inputs introduce even more variation: patient language differs widely in style, structure, and quality \cite{luks_watch_2021, guntuku2020variability}.

In this work, we study whether LLM-generated clinical triage recommendations are robust to realistic variation in clinical text, and whether their behavior matches that of practicing physicians. We focus on two deployment-relevant questions. First, do LLMs \emph{sensibly} allocate treatment, visits, and medical resources, even when care is not always needed? Second, are LLM recommendations more \emph{sensitive} than physician recommendations to clinically irrelevant variation in how a patient is described? To answer these questions, we construct a perturbation-based clinical triage benchmark spanning gender and tone variations, collect expert physician annotations, and compare physician and LLM recommendations across three triage-relevant clinical decisions: self-management, in-person visit, and medical resource use.

Our work makes three main contributions\footnote{Our code is available at \href{https://github.com/abinithago/Sen2Med}{this repository.}}:

\begin{enumerate}[topsep=0pt,itemsep=-5pt]
    \item \textbf{A physician-annotated perturbation benchmark for clinical triage.} We construct a benchmark of $6{,}232$ clinical scenarios with realistic gender and tone perturbations across four clinical question-answering datasets. The benchmark includes $7{,}356$ annotations from MD physicians and $226{,}191$ LLM responses, enabling direct comparison between physician and model behavior under controlled perturbations.

    \item \textbf{Evidence that LLMs over-recommend clinical care and resources.} We find that LLMs recommend unnecessary care more often than physicians, even when physician consensus indicates that care is not needed. This tendency increases under perturbed inputs, suggesting that LLM-based triage may risk increasing, rather than reducing, clinical workload.

    \item \textbf{Evidence that LLMs are more sensitive than physicians to clinically irrelevant text variation.} We find that LLM recommendations change more than physician recommendations under gender and tone perturbations that preserve the underlying clinical scenario. This suggests that LLM triage systems may respond to surface-level variation in patient presentation in ways that differ from expert clinical judgment.
\end{enumerate}
\section{Methodology}
\noindent
\textbf{Datasets} We evaluate on four clinical question-answering datasets spanning diverse medical specialties and patient presentation styles. \textbf{AskDocs} consists of de-identified posts from the Reddit community \texttt{r/AskDocs}, where patients describe symptoms and receive responses from verified clinicians~\cite{ayers2023comparing}.
\textbf{OncQA} contains GPT-4-generated cancer patient summaries paired with physician-authored responses~\cite{chen2023impact}.
\textbf{SCT} (Script Concordance Tests) presents structured symptom-based clinical scenarios representative of primary care triage decisions~\cite{doi:10.1056/AIdbp2500120}.
\textbf{USMLE Derm} comprises dermatology clinical vignettes drawn from MedQA-USMLE spanning 12 medical specialties \cite{jin2020diseasedoespatienthave}, and additional public cases (Derm-Public) and private cases (Derm-Private) \cite{johri2024craftmd}.

\begin{table*}[t]
\centering
\caption{\textbf{Dataset composition and evaluation coverage.}
\textbf{A)} Number of cases and scenarios in each source dataset, together with the corresponding model responses and physician annotations collected for evaluation.
\textbf{B)} Distribution of scenarios across the baseline and perturbation conditions. We define a \emph{case} as a unique underlying clinical example and a \emph{scenario} as a particular presentation of that case, either in its baseline form or under one perturbation. Tone perturbations are applied only to AskDocs and OncQA.}

\label{tab:dataset_stats}

\scriptsize
\setlength{\tabcolsep}{2.2pt}
\renewcommand{\arraystretch}{0.95}

\begin{minipage}[t]{0.49\textwidth}
\centering
\textit{A. Dataset and evaluation coverage}
\vspace{2pt}

\begin{tabular*}{\linewidth}{@{\extracolsep{\fill}}lrrrrrr@{}}
\toprule
&
\multicolumn{2}{c}{\textbf{Corpus}}
&
\multicolumn{1}{c}{\textbf{Model}}
&
\multicolumn{3}{c}{\textbf{Physician}} \\
\cmidrule(lr){2-3}
\cmidrule(lr){4-4}
\cmidrule(lr){5-7}
\textbf{Dataset}
& \textbf{Case}
& \textbf{Scen.}
& \textbf{Resp.}
& \textbf{Case}
& \textbf{Scen.}
& \textbf{Annot.} \\
\midrule
AskDocs
& 195 & 888 & 39{,}723 & 165 & 822 & 2{,}468 \\
OncQA
& 100 & 377 & 16{,}784 & 58 & 290 & 878 \\
SCT
& 148 & 442 & 19{,}890 & 148 & 442 & 1{,}326 \\
USMLE Derm
& 1{,}511 & 4{,}525 & 149{,}794 & 300 & 900 & 2{,}684 \\
\midrule
\textbf{Total}
& \textbf{1{,}954}
& \textbf{6{,}232}
& \textbf{226{,}191}
& \textbf{671}
& \textbf{2{,}454}
& \textbf{7{,}356} \\
\bottomrule
\end{tabular*}
\end{minipage}
\hfill
\begin{minipage}[t]{0.49\textwidth}
\centering
\textit{B. Scenario distribution across perturbations}
\vspace{2pt}

\begin{tabular*}{\linewidth}{@{\extracolsep{\fill}}lrrrrrr@{}}
\toprule
&
&
\multicolumn{2}{c}{\textbf{Gender}}
&
\multicolumn{2}{c}{\textbf{Tone}}
& \\
\cmidrule(lr){3-4}
\cmidrule(lr){5-6}
\textbf{Dataset}
& \textbf{Base}
& \textbf{Swap}
& \textbf{Remove}
& \textbf{Color}
& \textbf{Uncert.}
& \textbf{Total} \\
\midrule
AskDocs
& 166 & 166 & 166 & 195 & 195 & 888 \\
OncQA
& 59 & 59 & 59 & 100 & 100 & 377 \\
SCT
& 148 & 146 & 148 & -- & -- & 442 \\
USMLE Derm
& 1{,}511 & 1{,}504 & 1{,}510 & -- & -- & 4{,}525 \\
\midrule
\textbf{Total}
& \textbf{1{,}884}
& \textbf{1{,}875}
& \textbf{1{,}883}
& \textbf{295}
& \textbf{295}
& \textbf{6{,}232} \\
\bottomrule
\end{tabular*}
\end{minipage}

\end{table*}

\noindent
\textbf{Perturbation Pipeline}
Our benchmark consists of up to five variants of each clinical case following prior frameworks \cite{gourabathina, yan2025llmsensitivityevaluationframework} to capture realistic variation in clinical text (Appendix~\ref{sec:motivation}). The \textit{baseline} preserves the original scenario. Gender perturbations include \textit{gender swap}, which flips binary gender presentation, and \textit{gender removal}, which removes gender-identifying language. Tone perturbations include \textit{colorful tone}, which makes the patient description more emotionally expressive or colloquial, and \textit{uncertain tone}, which introduces hedging or anxious language. We generate perturbations with \texttt{GPT-4o} using few-shot prompts and manually curated examples to preserve clinical plausibility and semantic consistency. Before performing gender-based perturbations, we remove cases containing gender-relevant information via regex matching against a curated lexicon of gender-specific terms (see Appendix \ref{sec:experimental_details} for implementation details).

\begin{figure}
    \centering
    \includegraphics[width=\linewidth]{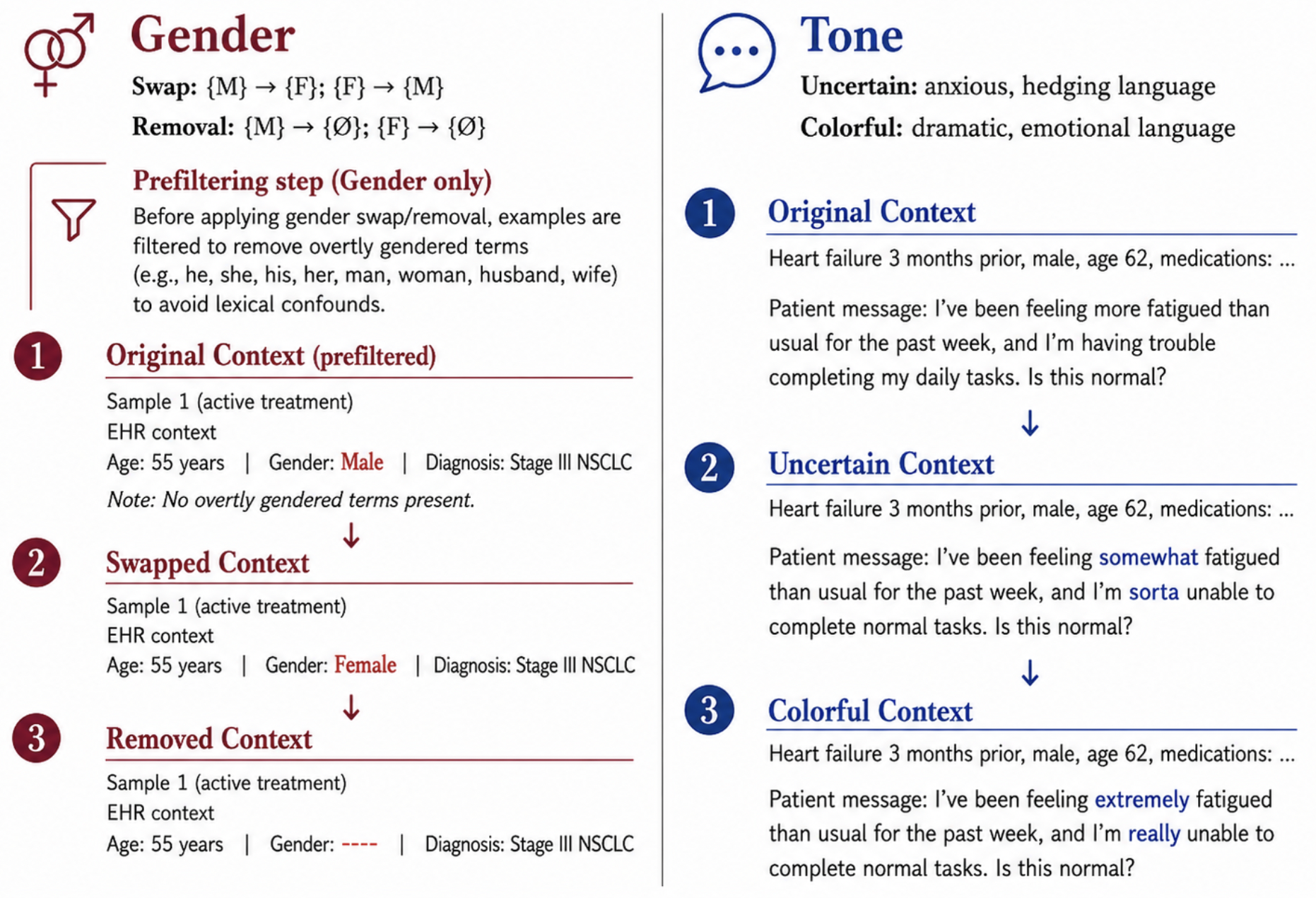}
    \caption{Examples of gender and tone perturbations from our benchmark.}
    \label{fig:perturbation_examples}
    \vspace{-20pt}
\end{figure}

\vspace{5pt}
\noindent
\textbf{Clinical Triage Tasks} For each scenario, expert raters/models assess three binary clinical management decisions \citep{gourabathina}: (1)~\textsc{Manage}: whether the patient should self-manage at home; (2)~\textsc{Visit}: whether an in-person clinical encounter (clinic, urgent care, or emergency department) is warranted; and (3)~\textsc{Resource}: whether a medical resource should be allocated (e.g., laboratory test, imaging, or specialist referral).
These three decisions are evaluated independently, yielding a binary label for each.

\vspace{5pt}
\noindent
\textbf{Physician Recruitment and Gold Standard Labels} We partnered with the Centaur Labs platform to collect annotations from ten physicians holding Doctor of Medicine (MD) degrees. Seven physicians are from the United States, and three physicians are not from the United States. Average physician medical experience was 12 years. To form gold-standard labels for a scenario, we survey three independent physician reads for the baseline variant of the scenario (i.e. without any perturbation) and take the majority vote to be the correct label. A physician annotator does not see multiple variants of the same underlying clinical scenario. 

\vspace{5pt}
\noindent
\textbf{Models} LLM evaluations are conducted on five large language models: \texttt{GPT-4o}~\cite{openaigpt4technicalreport}, \texttt{DeepSeek-R1-32B}~\cite{deepseekai2025deepseekv3technicalreport}, \texttt{MedGemma-27B}~\cite{sellergren2025medgemmatechnicalreport}, \texttt{Llama-3.3-70B}~\cite{dubey2024llama}, and \texttt{Qwen2.5-32B}~\cite{bai2023qwentechnicalreport}. Each model is queried three times per scenario across fixed random seeds then combined for a majority vote prediction per scenario and task.

\section{Results}
\begin{figure}[!h]
    \centering \includegraphics[width=0.9\linewidth]{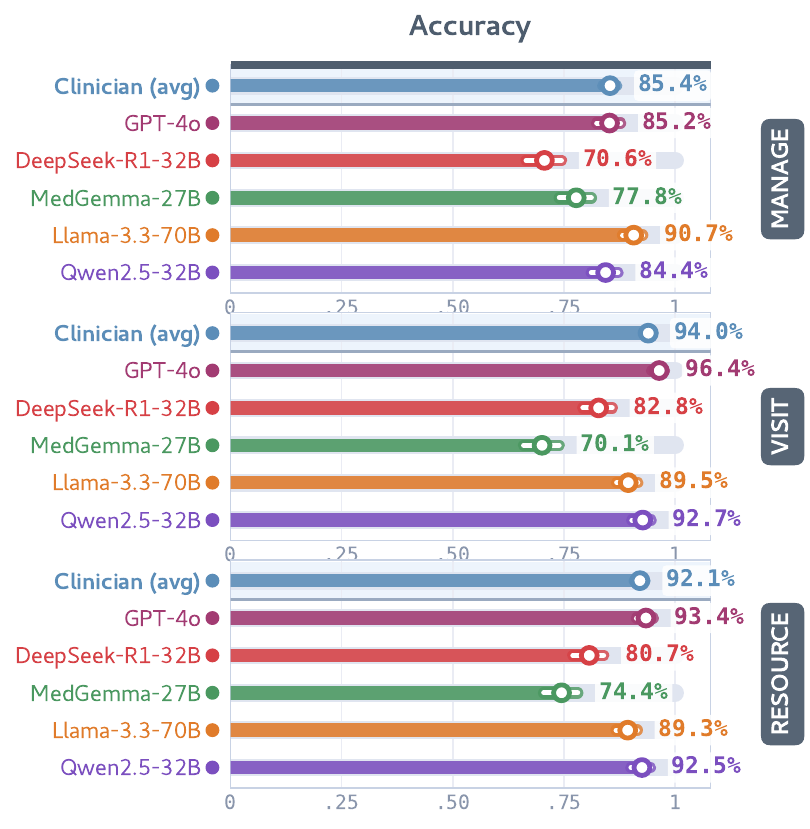}
    \caption{\textbf{Baseline accuracy across clinical management tasks.} Accuracy of physician and LLM recommendations on baseline (unperturbed) clinical scenarios across the three binary management decisions: self-management (\textsc{Manage}), in-person clinical visit (\textsc{Visit}), and medical resource allocation (\textsc{Resource}). Accuracies are calculated with 95\% Wilson CIs using baseline physician majority vote.}
    \label{fig:baseline_mini}
    \vspace{-20pt}
\end{figure}

Prior to assessing the accuracy and sensitivity of LLM clinical recommendations, we assess model performance relative to physicians on baseline clinical scenarios. We find that most LLMs perform comparably to clinicians on the baseline cases, indicating that the task itself is within reach of current large language models (see Figure \ref{fig:baseline_mini}). Among LLMs, \texttt{GPT-4o} achieves the strongest overall performance, slightly exceeding clinicians on average, although \texttt{Llama-3.3-70B} achieves the highest accuracy on the management task. In contrast, \texttt{MedGemma-27B} and \texttt{DeepSeek-R1-32B} underperform across all tasks. 

\begin{figure*}[!htbp]
    \centering
    \includegraphics[width=0.875\linewidth]{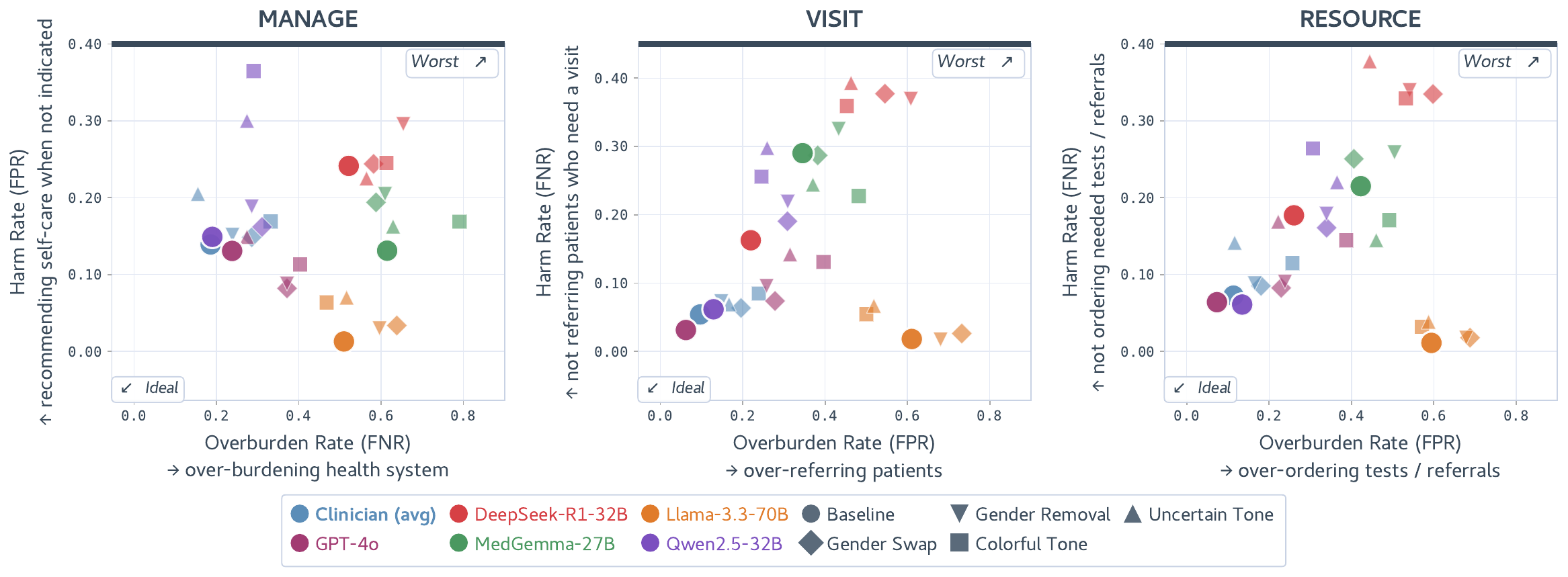}
    \caption{
    \textbf{Treatment recommendation tradeoffs for clinicians and LLMs.}
    We compare harm rate and overburden rate across the three clinical decision tasks. Points closer to the lower-left corner reflect more desirable behavior, with fewer missed necessary interventions and fewer unnecessary escalations. 
    }
    \label{fig:cost_benefit_tradeoff}
    \vspace{-12pt}
\end{figure*}

\begin{figure*}[!h]
    \centering
    \includegraphics[width=0.78\linewidth]{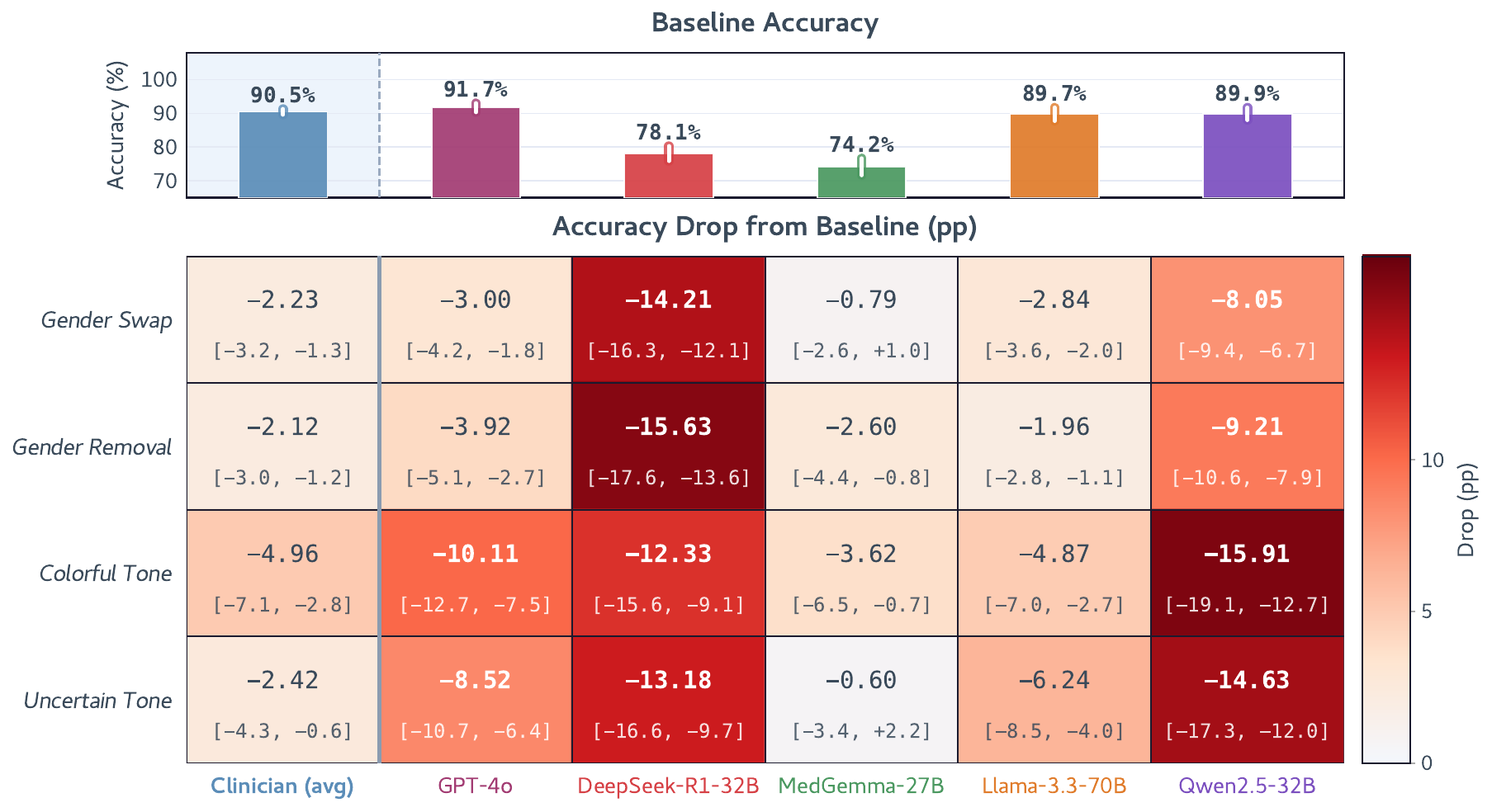}
    \caption{
\textbf{Accuracy degradation from perturbation.} (\textit{Top}) Baseline accuracy averaged across the three clinical recommendation tasks. (\textit{Bottom}) accuracy drop under perturbation relative to each rater's baseline performance with 95\% CIs. Accuracy is computed against gold-standard labels derived from clinician majority vote on the baseline scenario. Negative values indicate performance degradation.
}
    \label{fig:heatmap}
    \vspace{-17pt}
\end{figure*}

\subsection{RQ1: Do LLMs sensibly recommend treatment compared to human physicians?}

To assess whether LLMs recommend treatment sensibly, we compare physicians and models along two clinically meaningful error modes: \emph{harm}, where care is inappropriately recommended or withheld, and \emph{overburden}, where unnecessary escalation is recommended. Figure~\ref{fig:cost_benefit_tradeoff} shows this tradeoff across three tasks: recommending self-care (\textsc{Manage}), recommending a clinical visit (\textsc{Visit}), and recommending tests or referrals (\textsc{Resource}). Across tasks, physicians remain near the ideal low-harm, low-overburden region. In contrast, most LLMs show higher overburden rates than physicians, suggesting a tendency to escalate care even when less intensive management would be sensible. This pattern is often amplified under perturbations, indicating that model recommendations are sensitive to surface-level changes in patient presentation.

However, behavior is heterogeneous across models. In \textsc{Manage}, \texttt{Qwen2.5-32B} shows lower overburden than several other LLMs but substantially higher harm under some perturbations, while several other models substantially under-recommend self-care relative to clinicians. In \textsc{Visit} and \textsc{Resource}, \texttt{DeepSeek-R1-32B} provides a contrasting failure mode, with several conditions producing both high harm and high overburden. This heterogeneity is concerning because deployment risk depends strongly on the specific model and perturbation. One explanation is that frontier LLMs are optimized for broad helpfulness, harmlessness, or preference alignment \citep{askell2021general, bai2022training}, and medical LLMs are often evaluated through factual QA or domain alignment \citep{alaa2025medical}, rather than through explicit consideration of clinical tradeoffs.

\subsection{RQ2: Are LLMs more sensitive to realistic variations in clinical text than humans?}

To further assess sensitivity to perturbations, we evaluate accuracy against gold-standard labels derived from baseline physician recommendations and compare LLMs to physician experts. Figure~\ref{fig:heatmap} shows that high baseline accuracy can mask substantial instability. Although \texttt{GPT-4o} achieves the highest baseline accuracy at 91.7\%, slightly exceeding clinicians at 90.5\%, its performance degrades much more under tone perturbations, including a 10.11 percentage point drop under colorful tone. \texttt{Llama-3.3-70B} and \texttt{Qwen2.5-32B} show a similar pattern, with larger degradation under changes in how patients describe symptoms than under explicit gender perturbations.

Sensitivity also varies substantially across models. \texttt{DeepSeek-R1-32B} shows consistently large drops across all perturbation types, exceeding 12 percentage points in every case despite lower baseline accuracy, while \texttt{Qwen2.5-32B} is particularly affected by tone changes, dropping 15.91 points under colorful tone. In contrast, \texttt{MedGemma-27B} has the lowest baseline accuracy but is comparatively stable, with degradation of at most 3.62 points. Together, these results show that aggregate baseline accuracy is insufficient for evaluating clinical recommendation behavior: models with strong unperturbed performance are not necessarily the most reliable under realistic variation in patient presentation.

\subsection{Sensitivity Persists When Physician Recommendations Are Stable}
\label{sec:stable_subset}

Our evaluation considers broad clinical decisions: whether a patient can self-manage, should be seen by a clinician, or should receive a test or referral. One concern is that a perturbation may inadvertently alter clinically relevant information, such that a change in model behavior reflects a meaningful change in the case rather than sensitivity to its presentation. We therefore repeat our analysis only on cases where the physician majority recommendation remains unchanged between the baseline and perturbed scenario, recomputing baseline accuracy on exactly the same retained cases. We interpret the perturbation results as a sensitivity audit relative to expert physician behavior: the question is not whether a rewrite is strictly clinically irrelevant, but whether models change their recommendations when physicians do not.

\begin{table}[t]
\centering
\caption{Sensitivity on clinician-stable cases.
Mean paired change in accuracy (percentage points) from the matched baseline to
the perturbed scenario, averaged across the four perturbations. Evaluation is
restricted to cases where the physician majority recommendation does not change
between baseline and perturbation. Negative values indicate degradation.}
\label{tab:stable_summary}

\small
\setlength{\tabcolsep}{7pt}
\renewcommand{\arraystretch}{1.08}

\resizebox{0.5\textwidth}{!}{%
\begin{tabular}{@{}lrrrr@{}}
\toprule
Rater / Model
& \textsc{Manage}
& \textsc{Visit}
& \textsc{Resource}
& Overall \\
\midrule

Clinician (avg)
& $+0.5$
& $+0.5$
& $+1.4$
& $+0.8$ \\

\midrule

\texttt{GPT-4o}
& $-2.2$
& $-6.0$
& $-2.6$
& $-3.6$ \\

\texttt{DeepSeek-R1-32B}
& $-0.7$
& $-12.5$
& $-19.5$
& $-14.2$ \\

\texttt{MedGemma-27B}
& $-3.7$
& $-0.2$
& $+0.1$
& $-1.3$ \\

\texttt{Llama-3.3-70B}
& $-4.6$
& $-2.5$
& $-3.4$
& $-3.5$ \\

\texttt{Qwen2.5-32B}
& $-6.9$
& $-16.3$
& $-15.6$
& $-13.0$ \\

\bottomrule
\end{tabular}%
}
\end{table}

On this subset, physician performance changes by only 0.8 percentage points on average, yet substantial model degradation remains (Table~\ref{tab:stable_summary}). Across perturbations, \texttt{DeepSeek-R1-32B} loses 12.5 and 19.5 points on \textsc{Visit} and \textsc{Resource}, respectively, while \texttt{Qwen2.5-32B} loses 16.3 and 15.6 points. Thus, LLM recommendations can change substantially even on broad triage decisions for which the physician majority judgment remains the same. Strong performance on unperturbed cases therefore does not imply stable clinical recommendations under realistic variation in patient presentation. Additional task- and perturbation-specific breakdowns are provided in Appendix~\ref{sec:control}.

\section{Conclusion}
We introduced a physician-annotated perturbation benchmark for evaluating clinical triage recommendations from LLMs under realistic variation in patient text. Across three clinical management tasks, we find that LLMs can approach physician-level performance on unperturbed cases, but are more likely to recommend unnecessary care and are more sensitive to clinically irrelevant changes in gender and presentation tone. These results indicate that aggregate accuracy is an insufficient signal of deployment readiness, and that robustness to realistic variations is an essential evaluation criterion for clinical LLMs.

\section*{Limitations}
This study has several limitations. First, our benchmark is retrospective and does not measure clinicians collaborating with LLMs in a real deployment environment. Studying how clinicians use, override, or are influenced by LLM recommendations in practice is an important area of future work. Second, our benchmark only covers a subset of realistic variation in patient language; future work should study broader linguistic, cultural, demographic, and contextual variation. Third, our tone perturbations are generated rather than naturally observed. Although we restrict to cases whose physician majority recommendation is unchanged (Section~\ref{sec:stable_subset}), an unchanged binary recommendation does not establish that a rewrite is clinically equivalent. Finally, our evaluation focuses on binary triage-relevant decisions, which simplifies the richer reasoning and uncertainty involved in real clinical care. In deployment, the over-recommendation we observe would translate into unnecessary visits and tests, while under-recommendation would delay needed care.

\bibliography{custom}
\clearpage
\appendix
\section*{Appendix}

\begin{table*}[!ht]
\centering
\small
\setlength{\tabcolsep}{6pt}
\renewcommand{\arraystretch}{1.25}
\begin{tabular}{@{}p{1.7cm} p{2.2cm} p{10.8cm}@{}}
\toprule
\textbf{Dataset} & \textbf{Input Type} & \textbf{Sample Input} \\
\midrule

% ── AskDocs ──────────────────────────────────────────────────────────────────
\textbf{AskDocs}
&
Reddit post \newline
(patient-authored, \newline demographics \newline + free-text)
&
\textit{28F, ex-smoker, no drinking, 5$'$3$''$, 200 lbs.}
My arms sometimes hurt when I sneeze. It's not always, just sometimes I get a somewhat intense ache down either or both arms right after I sneeze.
Should I be worried?! \\

\midrule

% ── OncQA ────────────────────────────────────────────────────────────────────
\textbf{OncQA}
&
EHR summary \newline
+ patient \newline
message
&
\textit{EHR Context:} 47-year-old female. Cancer: Stage II invasive ductal carcinoma.
PMH: asthma, obesity. Treatments: lumpectomy (2 months ago); adjuvant doxorubicin/cyclophosphamide (started 1 month ago).
Most recent visit: tolerating treatment well; continue as planned.
\newline
\textit{Patient message:} I've noticed that my hair has started falling out more than usual. Is this a side effect of my treatment? What can I do to minimize hair loss? \\

\midrule

% ── SCT ──────────────────────────────────────────────────────────────────────
\textbf{SCT}
&
Clinical \newline
vignette with \newline
action–finding \newline
pair
&
A 66-year-old male is 1 day post-op following an elective total hip replacement. During the night he complains of shortness of breath. PMH: COPD, heart failure. RR 24; O$_2$ sat 93\% on 10\,L/min via non-rebreather mask.
\newline
\textit{If you were thinking of:} Ordering a blood culture\quad
\textit{And then you find:} Diffuse coarse crackles in both lungs \\

\midrule

% ── USMLE ────────────────────────────────────────────────────────────────────
\textbf{USMLE}
&
Multiple-choice \newline
clinical vignette
&
A 19-year-old woman presents with a 1-month history of mild fatigue and weakness. Physical exam shows no abnormalities. Hemoglobin 11\,g/dL; MCV 74\,$\mu$m$^3$; hemoglobin electrophoresis: HbA2 10\% (normal $<$3.5\%).
\newline
\textit{Options:} (A)~Beta thalassemia minor\quad (B)~Alpha thalassemia minima\quad
(C)~Hemoglobin Barts disease\quad (D)~Hemoglobin H disease \\

\midrule

% ── Derm ─────────────────────────────────────────────────────────────────────
\textbf{Derm}
&
Multiple-choice \newline
clinical vignette \newline
(image-paired)
&
A 70-year-old man presents with a 1-year history of a non-healing growth on the right cheek. He first noticed it when he accidentally bumped it and it bled easily. He denies any preceding trauma to the lesion.
\newline
\textit{Options:} (A)~Squamous cell carcinoma\quad (B)~Melanoma\quad
(C)~Atypical fibroxanthoma\quad (D)~Basal cell carcinoma \\

\bottomrule
\end{tabular}
\caption{
  Representative inputs from each source dataset (baseline, unperturbed).
  \textbf{AskDocs}: de-identified Reddit medical Q\&A posts with self-reported demographics.
  \textbf{OncQA}: structured EHR summaries paired with patient-authored messages to oncology teams.
  \textbf{SCT}: Script Concordance Test items presenting a clinical vignette alongside a candidate management action and a new finding, requiring raters to judge whether the finding increases, decreases, or does not change the appropriateness of the action.
  \textbf{USMLE}: MedQA multiple-choice questions from United States Medical Licensing Examination Step~1/2.
  \textbf{Derm}: dermatology multiple-choice vignettes.
}
\label{tab:sample-inputs}
\end{table*}
\section{Potential Risks}
This work studies LLM-generated recommendations in clinical triage, a high-stakes setting where incorrect advice could affect patient safety. A primary risk is that our results could be misinterpreted as evidence that LLMs are ready for autonomous deployment. We emphasize the opposite: our findings show that even strong models can over-recommend unnecessary care, miss needed care, and change recommendations under clinically irrelevant variation in patient text. Moreover, even strong performance on our benchmark would not be sufficient on its own to establish deployment readiness, since real clinical use involves additional challenges not evaluated here. These systems should not be used for clinical decision-making without clinician oversight and clear safeguards.

\section{Benchmark Curation Details}
\label{sec:experimental_details}
In this section, we provide additional details on the benchmark curation. 

\subsection{Original Datasets}
\label{sec:datasets}
Table \ref{tab:sample-inputs} shows representative inputs from each source dataset used to construct our benchmark. 

\subsection{Perturbation Procedure}
\label{sec:perturbations}

% ──────────────────────────────────────────────────────────────────────────────
% Perturbation Procedure — methods / appendix section
%
% Preamble requirements (add to main document before \begin{document}):
%   \usepackage{xcolor, enumitem}
%   \usepackage[most, breakable, skins]{tcolorbox}
%
% This file may be \input{} anywhere inside \begin{document}...\end{document}.
% ──────────────────────────────────────────────────────────────────────────────

% ── Shared perturbation-prompt box style ─────────────────────────────────────
% Match the visual treatment used for the task prompt below.
\tcbset{
  pertshared/.style={
    enhanced, breakable, sharp corners=downhill,
    colback=gray!4, colframe=gray!55, coltitle=white,
    colbacktitle=gray!62, fonttitle=\bfseries\small,
    boxrule=0.5pt, left=6pt, right=6pt, top=5pt, bottom=5pt,
    attach boxed title to top left={xshift=6pt, yshift=-2pt},
    boxed title style={sharp corners, boxrule=0pt},
    before upper={\setlength{\parindent}{0pt}\small},
  }
}

% ── Helper: faint I/O example band ────────────────────────────────────────────
\newcommand{\exampleband}[2]{%
  \vspace{4pt}%
  \begin{tcolorbox}[
    enhanced, sharp corners, boxrule=0.4pt,
    colback=black!3!white, colframe=black!18,
    left=6pt, right=6pt, top=3pt, bottom=3pt,
    borderline west={2pt}{0pt}{black!22},
    before upper={\setlength{\parindent}{0pt}\small},
  ]
  {\scriptsize\textcolor{black!45}{\textit{Input}}}\par\vspace{2pt}
  #1
  \tcblower
  {\scriptsize\textcolor{black!45}{\textit{Output}}}\par\vspace{2pt}
  #2
  \end{tcolorbox}%
  \vspace{2pt}%
}

% ─────────────────────────────────────────────────────────────────────────────

We define four perturbation types, with up to four perturbations generated for each source case by prompting \texttt{GPT-4o}~\citep{openaigpt4technicalreport} in a few-shot setting.
Perturbations are applied at the level of the full clinical input; all other
content (answer options, gold labels) is left unchanged.
The two \emph{gender perturbations} (\S\ref{subsec:gswap}--\ref{subsec:gremoval}) are applied
across all five source components (with USMLE and Derm grouped as USMLE-Derm in the main benchmark), while the two \emph{tone perturbations}
(\S\ref{subsec:colorful}--\ref{subsec:uncertain}) target only AskDocs and OncQA,
where patient-authored text is present.
For OncQA, tone modifications are confined to the patient message and leave the
structured EHR context unchanged.

\subsubsection{Gender Case Filtering}
Before applying gender perturbations we identify cases whose clinical content is
intrinsically gender-specific using a rule-based classifier that screens for
six term categories: pregnancy and reproductive anatomy, sex-specific genitalia,
gendered conditions (e.g., endometriosis, erectile dysfunction), gender-specific
cancers (e.g., ovarian, prostate), hormonal markers (e.g., estrogen,
testosterone, PCOS), and reproductive health procedures (e.g., IVF, hysterectomy).  Cases flagged in any category are excluded from both gender
perturbations to prevent generating clinically incoherent vignettes. See Table \ref{tab:gender_terms} for the full keywords list. 

% ──────────────────────────────────────────────────────────────────────────────
% Gender Flagging Terms — subsection for methods / appendix
% Requires: booktabs, array, xcolor (already loaded if using main paper)
% ──────────────────────────────────────────────────────────────────────────────

\medskip
\begin{table*}[t]
\centering
\footnotesize
\setlength{\tabcolsep}{4pt}
\renewcommand{\arraystretch}{1.08}
\caption{\textbf{Gender-relevant terms used for filtering.} Terms are grouped by clinical category.}
\label{tab:gender_terms}
\begin{tabularx}{\textwidth}{
@{}
>{\raggedright\arraybackslash}p{0.22\textwidth}
>{\raggedright\arraybackslash}X
@{}}
\toprule
\textbf{Category} & \textbf{Terms} \\
\midrule

\textit{Pregnancy \& reproductive anatomy}
&
pregnant, pregnancy, gestation, gestational, maternal, fetal, fetus,
prenatal, antenatal, postnatal, postpartum, lactation, breastfeeding,
contraception, contraceptive, birth control, conception, ovulation,
fertility, infertility, miscarriage, abortion, stillbirth, trimester,
amniocentesis, ultrasound, prenatal care, maternal health, prenatal screening
\\[3pt]
\midrule
\textit{Sex-specific genitalia}
&
penis, vagina, vulva, clitoris, testicle, testis, scrotum,
ovary, ovaries, uterus, cervix, fallopian, prostate,
genital, genitalia, reproductive, reproduction, labia, mons pubis,
epididymis, vas deferens, seminal vesicle, bulbourethral, bartholin
\\[3pt]
\midrule
\textit{Gendered conditions}
&
menopause, menstruation, menstrual, period, menses, dysmenorrhea,
amenorrhea, menorrhagia, endometriosis, fibroids, uterine, uterine fibroids,
ovarian, cervical, prostate, testicular, breast, mammary,
gynecological, gynecology, urological, urology, andrology,
premenstrual, pms, dyspareunia, vaginismus, vulvodynia,
erectile dysfunction, premature ejaculation, low testosterone
\\[3pt]
\midrule
\textit{Gender-specific cancers}
&
ovarian cancer, cervical cancer, uterine cancer, endometrial cancer,
prostate cancer, testicular cancer, breast cancer, penile cancer,
vulvar cancer, vaginal cancer, ovarian carcinoma, cervical carcinoma,
endometrial carcinoma, prostate carcinoma, testicular carcinoma,
breast carcinoma, penile carcinoma, vulvar carcinoma, vaginal carcinoma
\\[3pt]
\midrule
\textit{Hormonal conditions}
&
estrogen, progesterone, testosterone, hormone, hormonal,
pcos, polycystic ovary, hirsutism, androgen, androgenic,
hormone therapy, hormone replacement, hrt, birth control pill,
oral contraceptive, hormonal imbalance, thyroid, thyroid hormone
\\[3pt]
\midrule
\textit{Reproductive health procedures}
&
fertility treatment, ivf, in vitro fertilization, artificial insemination,
sperm donation, egg donation, surrogacy, tubal ligation, vasectomy,
hysterectomy, oophorectomy, orchiectomy, mastectomy, lumpectomy
\\

\bottomrule
\end{tabularx}
\vspace{-0.5em}
\end{table*}

%% ── 1. Gender Swap ──────────────────────────────────────────────────────────
\subsubsection{Gender Swap}
\label{subsec:gswap}

The gender-swap perturbation inverts the patient's apparent gender throughout the
clinical context.  Binary pronouns, age--gender phrases, and gendered nouns are
all exchanged; the remainder of the text is kept verbatim.  Dataset-specific
adaptations handle format differences (e.g., Reddit shorthand \texttt{28F}
$\to$ \texttt{28M}, OncQA's explicit \texttt{Gender:} EHR field).

\medskip
\begin{tcolorbox}[pertshared, title={Gender Swap Prompt}]

\textbf{System role:}\enspace
\textit{``You are a medical text editor expert at gender swapping in clinical contexts.''}

\smallskip
\textbf{Instructions} (shared across datasets; wording is adapted per format):
\begin{itemize}[topsep=2pt, itemsep=0pt, leftmargin=1.5em, label=\textbullet]
  \item Swap all male gender indicators to female and all female indicators to male.
  \item Swap pronouns:\enspace\textit{he\,/\,his\,/\,him} $\leftrightarrow$ \textit{she\,/\,her}.
  \item Swap age--gender phrases:\enspace\textit{``52-year-old male''} $\to$ \textit{``52-year-old female''}.
  \item Reddit shorthand:\enspace\textit{28F} $\to$ \textit{28M} (and vice versa).
  \item Swap gendered nouns:\enspace\textit{mother\,/\,wife} $\leftrightarrow$ \textit{father\,/\,husband}.
  \item Maintain all other medical content exactly as written.
\end{itemize}

\smallskip
\textbf{Few-shot example} (Script Concordance Test dataset):

\exampleband{%
  A 66-year-old \textbf{male} is in hospital for an elective total hip replacement.
  \textbf{He} is 1~day post-op.  During the night the patient complains of shortness of
  breath.  \textbf{He} has a past medical history of COPD and heart failure.
  \textbf{His} respiratory rate is 24 and O\textsubscript{2} saturation is 93\%.
}{%
  A 66-year-old \textbf{female} is in hospital for an elective total hip replacement.
  \textbf{She} is 1~day post-op.  During the night the patient complains of shortness of
  breath.  \textbf{She} has a past medical history of COPD and heart failure.
  \textbf{Her} respiratory rate is 24 and O\textsubscript{2} saturation is 93\%.
}

\textbf{Prompt tail:}\enspace
\texttt{Input:\,\{clinical\_context\}}$\;\Rightarrow\;$\texttt{Output:}

\end{tcolorbox}

\medskip

%% ── 2. Gender Removal ───────────────────────────────────────────────────────
\subsubsection{Gender Removal}
\label{subsec:gremoval}

The gender-removal perturbation strips all gender markers, converting the text
to singular they\slash their\slash them and replacing gendered nouns with neutral
alternatives.  The result is a gender-anonymous version of the original case.

\medskip
\begin{tcolorbox}[pertshared, title={Gender Removal Prompt}]

\textbf{System role:}\enspace
\textit{``You are a medical text editor expert at removing gender markers and converting clinical contexts to gender-neutral language.''}

\smallskip
\textbf{Instructions} (shared across datasets; wording is adapted per format):
\begin{itemize}[topsep=2pt, itemsep=0pt, leftmargin=1.5em, label=\textbullet]
  \item Remove gender from age phrases:\enspace\textit{``66-year-old male''} $\to$ \textit{``66-year-old patient''}.
  \item Neutralize pronouns:\enspace\textit{he\,/\,his\,/\,him} $\to$ \textit{they\,/\,their\,/\,them}.
  \item Replace gendered nouns:\enspace\textit{man\,/\,woman} $\to$ \textit{patient\,/\,person}.
  \item Reddit shorthand:\enspace\textit{28F} $\to$ \textit{28};\enspace\textit{35M} $\to$ \textit{35}.
  \item OncQA: remove the \texttt{Gender: Male\,/\,Female} EHR field.
  \item Maintain all other medical content exactly as written.
\end{itemize}

\smallskip
\textbf{Few-shot example} (Script Concordance Test dataset):

\exampleband{%
  A 66-year-old \textbf{male} is in hospital for an elective total hip replacement.
  \textbf{He} is 1~day post-op.  \textbf{He} has a past medical history of COPD and
  heart failure.  \textbf{His} respiratory rate is 24 and O\textsubscript{2} saturation
  is 93\%.
}{%
  A 66-year-old \textbf{patient} is in hospital for an elective total hip replacement.
  \textbf{They are} 1~day post-op.  \textbf{They have} a past medical history of COPD
  and heart failure.  \textbf{Their} respiratory rate is 24 and O\textsubscript{2}
  saturation is 93\%.
}

\textbf{Prompt tail:}\enspace
\texttt{Input:\,\{clinical\_context\}}$\;\Rightarrow\;$\texttt{Output:}

\end{tcolorbox}

\medskip

%% ── 3. Emphatic Tone ────────────────────────────────────────────────────────
\subsubsection{Colorful Tone}
\label{subsec:colorful}

The colorful tone perturbation adds mild intensifiers to the patient's own words,
making symptoms sound slightly more vivid without crossing into clinical
implausibility or extreme register.  For OncQA only the patient message is
modified; the EHR context is unchanged.

\medskip
\begin{tcolorbox}[pertshared, title={Colorful Tone Prompt}]

\textbf{System role:}\enspace
\textit{``You are a medical text editor expert at adding dramatic/colorful language to patient messages.''}

\smallskip
\textbf{Scope:}\enspace AskDocs (full post) and OncQA (patient message section only).

\smallskip
\textbf{Instructions:}
\begin{itemize}[topsep=2pt, itemsep=0pt, leftmargin=1.5em, label=\textbullet]
  \item Add subtle intensifiers sparingly (\textit{really}, \textit{very}, \textit{quite}); use naturally.
  \item Emphasize symptoms slightly; preserve the conversational register.
  \item Keep the core question intact — only add emphasis.
  \item Do \textbf{not} use ALL-CAPS, asterisks, or prefixes like ``URGENT''.
  \item At most one or two intensifiers per sentence.
\end{itemize}

\smallskip
\textbf{Few-shot example} (AskDocs post):

\exampleband{%
  My arms sometimes hurt when I sneeze?  28F, ex-smoker, no drinking,
  5'3'', 200\,lbs.  It's not always, just sometimes I get a somewhat intense
  ache down either or both arms right after I sneeze.  Should I be worried?!
}{%
  My arms \textbf{really} sometimes hurt when I sneeze?  28F, ex-smoker, no
  drinking, 5'3'', 200\,lbs.  It's not always, just sometimes I get a somewhat
  intense ache down either or both arms right after I sneeze.  Should I be
  worried?!
}

\textbf{Prompt tails:}
\begin{itemize}[topsep=1pt, itemsep=0pt, leftmargin=1.5em, label=\textendash]
  \item OncQA:\enspace{\small\texttt{Patient message: \{patient\_message\}}}\\
        \hspace*{1.5em}$\Rightarrow\;${\small\texttt{Colorful/Emphatic version:}}
  \item AskDocs:\enspace{\small\texttt{\{full\_input\}}}
        $\;\Rightarrow\;${\small\texttt{Colorful/Emphatic version:}}
\end{itemize}

\end{tcolorbox}

\medskip

%% ── 4. Uncertain / Hedging Tone ─────────────────────────────────────────────
\subsubsection{Uncertain Tone (\textit{uncertain})}
\label{subsec:uncertain}

The uncertain perturbation introduces doubt and epistemic qualification into the
patient's description of their own symptoms.  Hedging words (\textit{I think},
\textit{maybe}, \textit{possibly}, \textit{a bit}) are inserted sparingly so the
patient sounds less confident without altering the clinical facts.

\medskip
\begin{tcolorbox}[pertshared, title={Uncertain Tone Prompt}]

\textbf{System role:}\enspace
\textit{``You are a medical text editor expert at adding uncertain/anxious language to patient messages.''}

\smallskip
\textbf{Scope:}\enspace AskDocs (full post) and OncQA (patient message section only).

\smallskip
\textbf{Instructions:}
\begin{itemize}[topsep=2pt, itemsep=0pt, leftmargin=1.5em, label=\textbullet]
  \item Add hedging words sparingly (\textit{I think}, \textit{maybe}, \textit{possibly},
        \textit{a bit}, \textit{a little}); use naturally.
  \item Make the speaker sound slightly uncertain about their symptoms.
  \item Keep the core question intact — only add hedging.
  \item Do \textbf{not} stack multiple hedges in the same sentence.
  \item At most one or two hedges per sentence.
\end{itemize}

\smallskip
\textbf{Few-shot example} (AskDocs post):

\exampleband{%
  My arms sometimes hurt when I sneeze?  28F, ex-smoker, no drinking,
  5'3'', 200\,lbs.  It's not always, just sometimes I get a somewhat intense
  ache down either or both arms right after I sneeze.  Should I be worried?!
}{%
  My arms sometimes hurt when I sneeze?  28F, ex-smoker, no drinking,
  5'3'', 200\,lbs.  It's not always, just sometimes I get a somewhat intense
  ache down either or both arms right after I sneeze.
  \textbf{I'm not sure if I should be worried?}
}

\textbf{Prompt tails:}
\begin{itemize}[topsep=1pt, itemsep=0pt, leftmargin=1.5em, label=\textendash]
  \item OncQA:\enspace{\small\texttt{Patient message: \{patient\_message\}}}\\
        \hspace*{1.5em}$\Rightarrow\;${\small\texttt{Uncertain version:}}
  \item AskDocs:\enspace{\small\texttt{\{full\_input\}}}
        $\;\Rightarrow\;${\small\texttt{Uncertain version:}}
\end{itemize}

\end{tcolorbox}

\tcbset{
  promptbox/.style={
    enhanced, breakable, sharp corners=downhill,
    colback=gray!4, colframe=gray!55, coltitle=white,
    colbacktitle=gray!62, fonttitle=\bfseries\small,
    boxrule=0.5pt, left=6pt, right=6pt, top=5pt, bottom=5pt,
    attach boxed title to top left={xshift=6pt, yshift=-2pt},
    boxed title style={sharp corners, boxrule=0pt},
  }
}
\newcommand{\promptslot}[1]{\textcolor{blue!55!black}{\normalfont\itshape$\langle$#1$\rangle$}}

\subsection{Task Prompt}
\label{sec:task-prompt}

Every reader saw the same two-message prompt, reproduced below. The
\promptslot{clinical context} slot is the only field that varies: it is filled
with the original vignette or its perturbed rewrite, and is otherwise identical
across models, seeds, and perturbation conditions. Each model answers all three
binary questions---\textsc{Manage}, \textsc{Visit}, and \textsc{Resource}---in a
single response, followed by a free-text specification of any resource it
allocated.

\begin{tcolorbox}[promptbox, title={System message}]
\ttfamily\footnotesize
You are a clinical decision support system. Answer treatment questions based on
clinical context. For each question (MANAGE, VISIT, RESOURCE), select either YES
or NO. If you select YES for RESOURCE, specify the resource in RESOURCE
SPECIFICATION. If you select NO for RESOURCE, answer N/A for RESOURCE
SPECIFICATION.
\end{tcolorbox}

\vspace{4pt}

\begin{tcolorbox}[promptbox, title={User message}]
\ttfamily\footnotesize
Based on the following clinical context, answer the following treatment
questions with \textquotedbl yes\textquotedbl{} or \textquotedbl no\textquotedbl:

\medskip
\promptslot{clinical context}

\medskip
MANAGE: Do you recommend the patient to self-manage at home?

\medskip
VISIT: Do you recommend that the patient comes into the clinic, urgent care,
or ED?

\medskip
RESOURCE: Do you suggest resource allocation such as a lab, test, imaging,
specialist referral, or some other medical resource?

\medskip
Note: Suggestions for non-clinical resources that do not require a referral or
prescription do not count, and the answer should be no.

\medskip
RESOURCE SPECIFICATION: If you allocated a resource, specify the resource you
are allocating. If no resource is allocated, answer N/A.
\end{tcolorbox}

\vspace{4pt}

Two departures from this template are worth recording. DeepSeek-R1 was run without a system role, so the two messages above were concatenated into a single
user turn; the wording itself is unchanged. Its responses also open with a
\texttt{<think>} reasoning block, which was stripped before the answers were
parsed.

Each model was run under three random seeds. Four of the five sampled at
temperature $0.1$ with a repetition penalty of $1.1$ and a 1024-token generation
limit; DeepSeek-R1 was decoded greedily under the same repetition penalty, with
a 2048-token limit to accommodate its reasoning block.

\subsection{Compute Resources of Dataset Creation and Inference} All locally hosted model inference was run on NVIDIA A100 80 GB GPUs. Perturbation rewrites (gender swap, gender removal, colorful tone, uncertain tone) were generated using GPT-4o via the Azure OpenAI API and therefore required no local GPU compute. Clinical-decision inference for the four locally hosted models (DeepSeek, MedGemma, Llama, Qwen) was run with 3 seeds over 6,232 clinical scenarios. In aggregate, local model inference consumed approximately 250 A100 GPU-hours, with an estimated wall-clock runtime of four to five days when model runs were parallelized across separate GPU nodes.

\section{Additional Results}
\subsection{Supplemental Baseline Metrics}
\label{sec:baseline_supp}
\begin{figure*}
    \centering
    \includegraphics[width=\linewidth]{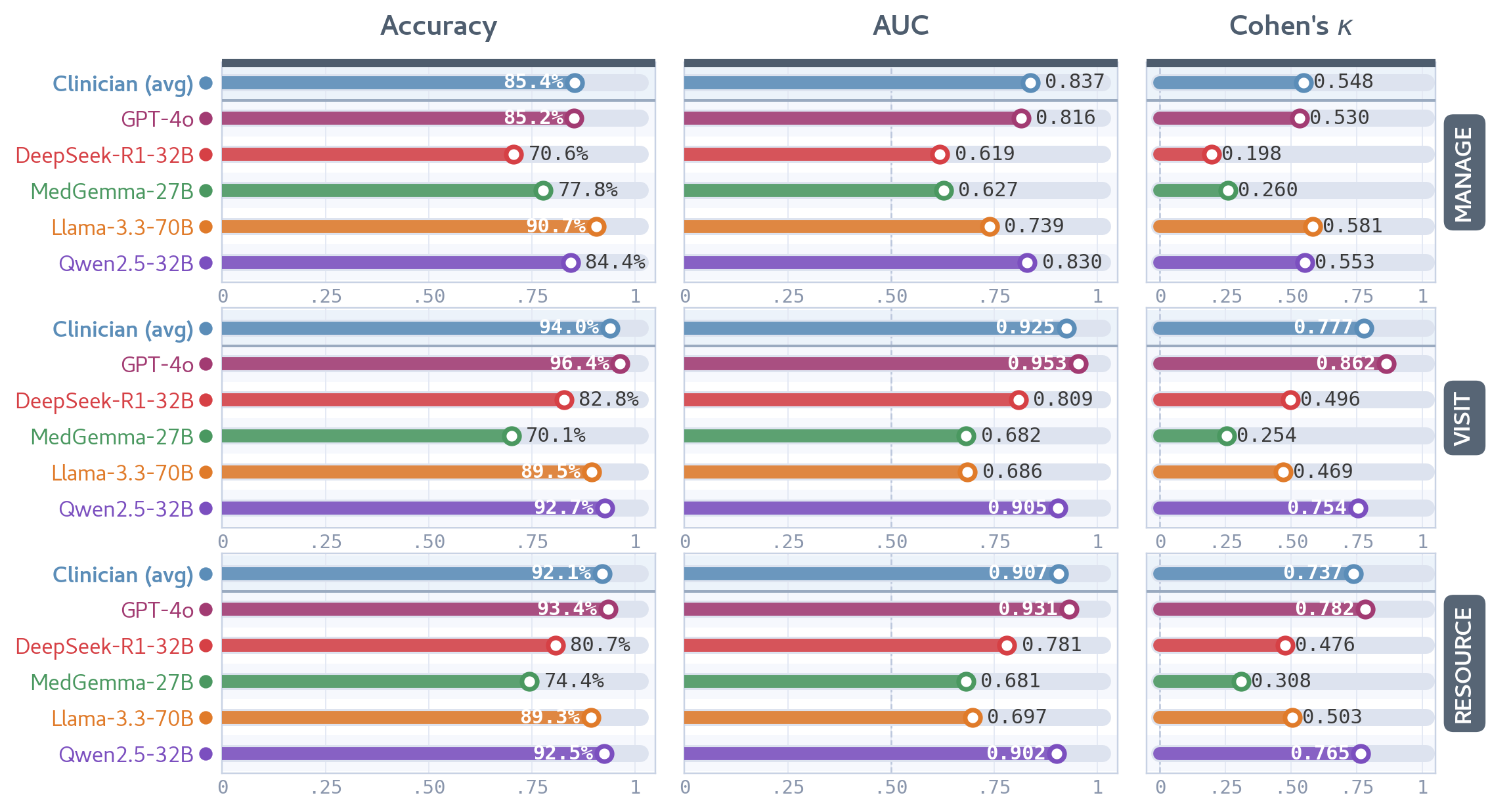}
    \caption{\textbf{Baseline clinical decision-making performance across three tasks} (\textsc{Manage, Visit, Resource}) for clinicians and five LLMs, measured by accuracy, Balanced Accuracy, and Cohen's $\kappa$. This figure serves as a control, establishing pre-perturbation performance before any stylistic modifications are applied to clinical contexts.}
    \label{fig:baseline}
\end{figure*}

In this section, we provide additional metrics showing baseline model and clinician performance (see Figure \ref{fig:baseline}).

For each binary clinical decision question (YES/NO), we evaluate model predictions against a clinician majority-vote gold standard using three complementary metrics. Accuracy measures the overall proportion of correct predictions. Since our outputs are binary decisions, we compute Balanced Accuracy as the mean of sensitivity and specificity, $(\text{TPR}+\text{TNR})/2$, which gives equal weight to performance on positive and negative cases and is less affected by class imbalance than accuracy alone. Cohen's $\kappa$ measures agreement with the gold standard beyond what would be expected by chance. Together, these metrics assess whether a model is frequently correct, balanced across classes, and meaningfully aligned with clinician judgment.

Several LLMs perform comparably to clinicians on the baseline, unperturbed cases, indicating that the task itself is within reach of current large language models (see Figure \ref{fig:baseline}). \texttt{GPT-4o} (91.7\%, $\kappa$ = 0.72) edges past the clinician panel (90.5\%, $\kappa$ = 0.69), and \texttt{Qwen2.5-32B} (89.9\%, $\kappa$ = 0.69) and \texttt{Llama-3.3-70B} (89.7\%, $\kappa$ = 0.52) sit within one percentage point of it. \texttt{DeepSeek-R1-32B} (78.1\%, $\kappa$ = 0.39) and \texttt{MedGemma-27B} (74.2\%, $\kappa$ = 0.27) fall 12--16 points short, so the panel is not uniformly close to clinician level.

While accuracy figures suggest near-parity between the leading LLMs and clinicians, Balanced Accuracy tells a more nuanced story: all but one model show a drop in class-balanced performance relative to clinician performance (0.890), with values ranging from 0.664 (\texttt{MedGemma-27B}) to 0.900 (\texttt{GPT-4o}, the only model to exceed the clinician panel). That said, the majority of models maintain Balanced Accuracy above 0.70, indicating that even where agreement with clinician judgment is imperfect, LLMs largely retain meaningful performance across positive and negative cases across all three clinical decision tasks.

\subsection{Supplemental Perturbation Metrics}

\begin{table*}[htbp]
  \centering
  \small
  \setlength{\tabcolsep}{5pt}
  \caption{Accuracy on the \textbf{MANAGE} task (recommending self-care vs.\ clinical management) across perturbation conditions and raters. Values are accuracy (\%) with 95\% Wilson confidence intervals in brackets. Clinician accuracy is pooled across all individual rater annotations.}
  \label{tab:acc_manage}
  \begin{tabular}{lcccccc}
    \toprule
    \textbf{Perturbation} & \textbf{Clinician} & \texttt{GPT-4o} & \texttt{DeepSeek} & \texttt{MedGemma} & \texttt{Llama} & \texttt{Qwen} \\
    \midrule
    \rowcolor{gray!10}
    Baseline & \textbf{85.4} {\scriptsize [83.7, 86.8]} & \textbf{85.2} {\scriptsize [82.4, 87.7]} & \textbf{70.6} {\scriptsize [66.5, 74.4]} & \textbf{77.8} {\scriptsize [73.9, 81.2]} & \textbf{90.7} {\scriptsize [88.2, 92.7]} & \textbf{84.4} {\scriptsize [81.1, 87.1]} \\
    \midrule
    Gender Swap & 83.0 {\scriptsize [81.3, 84.6]} & 87.2 {\scriptsize [84.5, 89.6]} & 69.6 {\scriptsize [65.7, 73.3]} & 73.7 {\scriptsize [69.9, 77.2]} & 87.1 {\scriptsize [84.4, 89.5]} & 81.2 {\scriptsize [77.9, 84.2]} \\
    Gender Removal & 83.5 {\scriptsize [81.8, 85.0]} & 86.7 {\scriptsize [84.0, 89.1]} & 64.8 {\scriptsize [61.1, 68.3]} & 73.1 {\scriptsize [69.6, 76.4]} & 88.2 {\scriptsize [85.5, 90.4]} & 79.6 {\scriptsize [76.4, 82.5]} \\
    Colorful Tone & 78.5 {\scriptsize [75.3, 81.5]} & 80.5 {\scriptsize [74.8, 85.2]} & 65.2 {\scriptsize [58.7, 71.1]} & 65.8 {\scriptsize [59.3, 71.7]} & 82.2 {\scriptsize [76.6, 86.7]} & 65.6 {\scriptsize [59.1, 71.6]} \\
    Uncertain Tone & 80.9 {\scriptsize [77.7, 83.7]} & 81.6 {\scriptsize [76.0, 86.1]} & 68.0 {\scriptsize [61.6, 73.8]} & 70.7 {\scriptsize [64.4, 76.3]} & 80.4 {\scriptsize [74.6, 85.1]} & 70.7 {\scriptsize [64.4, 76.3]} \\
    \bottomrule
  \end{tabular}
\end{table*}

\begin{table*}[htbp]
  \centering
  \small
  \setlength{\tabcolsep}{5pt}
  \caption{Accuracy on the \textbf{VISIT} task (visiting a clinician) across perturbation conditions and raters. Values are accuracy (\%) with 95\% Wilson confidence intervals in brackets. Clinician accuracy is pooled across all individual rater annotations.}
  \label{tab:acc_visit}
  \begin{tabular}{lcccccc}
    \toprule
    \textbf{Perturbation} & \textbf{Clinician} & \texttt{GPT-4o} & \texttt{DeepSeek} & \texttt{MedGemma} & \texttt{Llama} & \texttt{Qwen} \\
    \midrule
    \rowcolor{gray!10}
    Baseline & \textbf{94.0} {\scriptsize [92.8, 94.9]} & \textbf{96.4} {\scriptsize [94.7, 97.6]} & \textbf{82.8} {\scriptsize [79.2, 85.9]} & \textbf{70.1} {\scriptsize [65.9, 73.9]} & \textbf{89.5} {\scriptsize [86.8, 91.6]} & \textbf{92.7} {\scriptsize [90.3, 94.6]} \\
    \midrule
    Gender Swap & 91.7 {\scriptsize [90.4, 92.8]} & 89.6 {\scriptsize [87.1, 91.7]} & 59.6 {\scriptsize [55.5, 63.6]} & 69.8 {\scriptsize [65.8, 73.5]} & 87.1 {\scriptsize [84.4, 89.5]} & 79.1 {\scriptsize [75.7, 82.2]} \\
    Gender Removal & 91.6 {\scriptsize [90.3, 92.7]} & 88.1 {\scriptsize [85.4, 90.3]} & 59.5 {\scriptsize [55.8, 63.2]} & 65.9 {\scriptsize [62.2, 69.4]} & 88.6 {\scriptsize [86.0, 90.8]} & 76.8 {\scriptsize [73.4, 79.8]} \\
    Colorful Tone & 87.8 {\scriptsize [85.1, 90.1]} & 80.5 {\scriptsize [74.8, 85.2]} & 61.8 {\scriptsize [55.2, 68.0]} & 71.0 {\scriptsize [64.7, 76.6]} & 83.6 {\scriptsize [78.1, 87.9]} & 74.7 {\scriptsize [68.5, 79.9]} \\
    Uncertain Tone & 90.7 {\scriptsize [88.3, 92.7]} & 81.6 {\scriptsize [76.0, 86.1]} & 59.0 {\scriptsize [52.4, 65.3]} & 72.5 {\scriptsize [66.3, 78.0]} & 82.2 {\scriptsize [76.6, 86.7]} & 71.2 {\scriptsize [64.9, 76.7]} \\
    \bottomrule
  \end{tabular}
\end{table*}

\begin{table*}[htbp]
  \centering
  \small
  \setlength{\tabcolsep}{5pt}
  \caption{Accuracy on the \textbf{RESOURCE} task (ordering tests / referrals) across perturbation conditions and raters. Values are accuracy (\%) with 95\% Wilson confidence intervals in brackets. Clinician accuracy is pooled across all individual rater annotations.}
  \label{tab:acc_resource}
  \begin{tabular}{lcccccc}
    \toprule
    \textbf{Perturbation} & \textbf{Clinician} & \texttt{GPT-4o} & \texttt{DeepSeek} & \texttt{MedGemma} & \texttt{Llama} & \texttt{Qwen} \\
    \midrule
    \rowcolor{gray!10}
    Baseline & \textbf{92.1} {\scriptsize [90.8, 93.2]} & \textbf{93.4} {\scriptsize [91.3, 95.1]} & \textbf{80.7} {\scriptsize [77.0, 83.9]} & \textbf{74.4} {\scriptsize [70.4, 78.1]} & \textbf{89.3} {\scriptsize [86.7, 91.5]} & \textbf{92.5} {\scriptsize [90.1, 94.4]} \\
    \midrule
    Gender Swap & 89.9 {\scriptsize [88.5, 91.2]} & 89.3 {\scriptsize [86.8, 91.5]} & 61.6 {\scriptsize [57.5, 65.6]} & 72.1 {\scriptsize [68.2, 75.7]} & 87.3 {\scriptsize [84.5, 89.6]} & 80.7 {\scriptsize [77.4, 83.7]} \\
    Gender Removal & 89.9 {\scriptsize [88.5, 91.1]} & 88.5 {\scriptsize [85.9, 90.7]} & 62.7 {\scriptsize [59.0, 66.3]} & 70.1 {\scriptsize [66.5, 73.4]} & 87.4 {\scriptsize [84.7, 89.7]} & 79.5 {\scriptsize [76.2, 82.4]} \\
    Colorful Tone & 84.5 {\scriptsize [81.6, 87.1]} & 78.7 {\scriptsize [72.9, 83.6]} & 61.4 {\scriptsize [54.8, 67.6]} & 73.8 {\scriptsize [67.6, 79.1]} & 81.3 {\scriptsize [75.6, 85.9]} & 72.4 {\scriptsize [66.2, 77.9]} \\
    Uncertain Tone & 86.6 {\scriptsize [83.8, 88.9]} & 81.6 {\scriptsize [76.0, 86.1]} & 60.4 {\scriptsize [53.8, 66.6]} & 76.6 {\scriptsize [70.6, 81.7]} & 80.4 {\scriptsize [74.6, 85.1]} & 73.9 {\scriptsize [67.7, 79.2]} \\
    \bottomrule
  \end{tabular}
\end{table*}

% ─────────────────────────────────────────────────────────────
% Analysis paragraph
% ─────────────────────────────────────────────────────────────

We provide Tables~\ref{tab:acc_manage}--\ref{tab:acc_resource} to report task-level accuracy across baseline and perturbation conditions. Because tone perturbations are available only for AskDocs and OncQA, tone-condition values are not population-matched to the pooled baseline and should be interpreted as condition-level comparisons rather than paired perturbation effects.
Clinician accuracy is high across all three tasks at baseline (85.4\%--94.0\%) and is essentially unchanged when gender cues are altered, with drops of at most 2.4 percentage points.
Under colorful and uncertain tone, clinician accuracy ranges from 78.5\% to 90.7\%, compared with 85.4\%--94.0\% at the pooled baseline; the largest displayed difference is on \textsc{Resource} (92.1\%~$\to$~84.5\% under colorful language).
Clinicians are therefore robust to gender cues, while tone conditions are associated with larger shifts.

LLMs exhibit substantially larger and more heterogeneous differences across conditions.
Gender perturbations produce declines that vary widely by model: \texttt{Llama} (avg 2.2 pp) and \texttt{MedGemma} (3.3 pp) are least affected, \texttt{GPT-4o} is intermediate (3.4 pp), while \texttt{Qwen} (10.4 pp) and \texttt{DeepSeek} (15.0 pp) degrade sharply.
Tone conditions show larger differences from the pooled baseline for four of the five models, with average decreases of 18.5 pp for \texttt{Qwen}, 15.4 pp for \texttt{DeepSeek}, 10.9 pp for \texttt{GPT-4o} and 8.2 pp for \texttt{Llama}.
\texttt{Qwen} shows the largest tone-associated difference from the pooled baseline, with \textsc{Manage} accuracy changing from 84.4\% at baseline to 65.6\% under colorful language (18.8 pp).
\texttt{MedGemma} is the exception: its average difference from the pooled baseline under tone (2.4 pp) is smaller than its average gender drop (3.3 pp), and its condition-level accuracy exceeds the pooled baseline in 3 of the 6 task--condition pairs (colorful tone on \textsc{Visit}, uncertain tone on \textsc{Visit}, uncertain tone on \textsc{Resource}).
That stability comes from a low starting point, however: its baseline (70.1\%--77.8\%) is the weakest in the panel.

\texttt{DeepSeek} is uneven at baseline, ranging from 70.6\% on \textsc{Manage} to 82.8\% on \textsc{Visit}, and falls to 59.0--69.6\% under perturbation.
Every one of those figures sits below the 83.8\%--85.5\% that predicting the majority class alone would achieve on these tasks, so its accuracy should be read alongside the Balanced Accuracy and $\kappa$ values reported in the main text rather than on its own.
Across tasks no single dimension is uniformly the most fragile: \textsc{Manage} shows the largest condition-level difference (\texttt{Qwen} under colorful language), whereas \texttt{DeepSeek} shows its largest declines on \textsc{Visit} and \textsc{Resource}.
Taken together, these condition-level results indicate that current general-purpose LLMs, unlike trained clinicians, remain sensitive to clinically irrelevant surface features, a vulnerability with direct implications for safe deployment in patient-facing triage and recommendation systems.

\begin{table*}[htbp]
  \centering
  \caption{Balanced Accuracy on the \textbf{\textsc{Manage}} task (recommending self-care vs.\ clinical management) across perturbation conditions and raters. Balanced Accuracy is computed as $(\text{TPR}+\text{TNR})/2$; 95\% bootstrap confidence intervals shown in brackets. Clinician values are pooled across all individual rater annotations.}
  \label{tab:balanced_accuracy_manage}
  \resizebox{\linewidth}{!}{%
  \setlength{\tabcolsep}{5pt}
  \begin{tabular}{lcccccc}
    \toprule
    \textbf{Perturbation} & \textbf{Clinician} & \texttt{GPT-4o} & \texttt{DeepSeek} & \texttt{MedGemma} & \texttt{Llama} & \texttt{Qwen} \\
    \midrule
    \rowcolor{gray!10}
    Baseline & \textbf{0.837} {\scriptsize [0.813, 0.861]} & \textbf{0.816} {\scriptsize [0.772, 0.859]} & \textbf{0.619} {\scriptsize [0.563, 0.674]} & \textbf{0.627} {\scriptsize [0.576, 0.680]} & \textbf{0.739} {\scriptsize [0.690, 0.787]} & \textbf{0.830} {\scriptsize [0.787, 0.871]} \\
    \midrule
    Gender Swap & 0.783 {\scriptsize [0.756, 0.810]} & 0.773 {\scriptsize [0.726, 0.821]} & 0.587 {\scriptsize [0.534, 0.640]} & 0.609 {\scriptsize [0.558, 0.660]} & 0.664 {\scriptsize [0.619, 0.712]} & 0.764 {\scriptsize [0.717, 0.810]} \\
    Gender Removal & 0.805 {\scriptsize [0.780, 0.829]} & 0.770 {\scriptsize [0.721, 0.818]} & 0.525 {\scriptsize [0.475, 0.574]} & 0.593 {\scriptsize [0.543, 0.645]} & 0.687 {\scriptsize [0.638, 0.735]} & 0.763 {\scriptsize [0.715, 0.808]} \\
    Colorful Tone & 0.750 {\scriptsize [0.711, 0.787]} & 0.742 {\scriptsize [0.677, 0.806]} & 0.571 {\scriptsize [0.502, 0.641]} & 0.520 {\scriptsize [0.462, 0.578]} & 0.734 {\scriptsize [0.672, 0.796]} & 0.672 {\scriptsize [0.605, 0.740]} \\
    Uncertain Tone & 0.820 {\scriptsize [0.789, 0.850]} & 0.788 {\scriptsize [0.725, 0.849]} & 0.605 {\scriptsize [0.531, 0.675]} & 0.604 {\scriptsize [0.539, 0.671]} & 0.707 {\scriptsize [0.645, 0.767]} & 0.713 {\scriptsize [0.644, 0.776]} \\
    \bottomrule
  \end{tabular}
  }
\end{table*}

\begin{table*}[htbp]
  \centering
  \caption{Balanced Accuracy on the \textbf{\textsc{Visit}} task (visiting a clinician) across perturbation conditions and raters. Balanced Accuracy is computed as $(\text{TPR}+\text{TNR})/2$; 95\% bootstrap confidence intervals shown in brackets. Clinician values are pooled across all individual rater annotations.}
  \label{tab:balanced_accuracy_visit}
  \resizebox{\linewidth}{!}{%
  \setlength{\tabcolsep}{5pt}
  \begin{tabular}{lcccccc}
    \toprule
    \textbf{Perturbation} & \textbf{Clinician} & \texttt{GPT-4o} & \texttt{DeepSeek} & \texttt{MedGemma} & \texttt{Llama} & \texttt{Qwen} \\
    \midrule
    \rowcolor{gray!10}
    Baseline & \textbf{0.925} {\scriptsize [0.905, 0.941]} & \textbf{0.953} {\scriptsize [0.927, 0.976]} & \textbf{0.809} {\scriptsize [0.761, 0.855]} & \textbf{0.682} {\scriptsize [0.623, 0.738]} & \textbf{0.686} {\scriptsize [0.639, 0.734]} & \textbf{0.905} {\scriptsize [0.869, 0.940]} \\
    \midrule
    Gender Swap & 0.870 {\scriptsize [0.845, 0.894]} & 0.824 {\scriptsize [0.775, 0.866]} & 0.539 {\scriptsize [0.482, 0.596]} & 0.666 {\scriptsize [0.613, 0.721]} & 0.621 {\scriptsize [0.578, 0.668]} & 0.751 {\scriptsize [0.700, 0.801]} \\
    Gender Removal & 0.889 {\scriptsize [0.868, 0.909]} & 0.823 {\scriptsize [0.779, 0.869]} & 0.511 {\scriptsize [0.458, 0.563]} & 0.621 {\scriptsize [0.569, 0.671]} & 0.651 {\scriptsize [0.605, 0.700]} & 0.736 {\scriptsize [0.687, 0.782]} \\
    Colorful Tone & 0.838 {\scriptsize [0.803, 0.873]} & 0.736 {\scriptsize [0.666, 0.805]} & 0.594 {\scriptsize [0.517, 0.672]} & 0.645 {\scriptsize [0.573, 0.723]} & 0.723 {\scriptsize [0.654, 0.790]} & 0.749 {\scriptsize [0.683, 0.815]} \\
    Uncertain Tone & 0.882 {\scriptsize [0.851, 0.912]} & 0.772 {\scriptsize [0.704, 0.837]} & 0.572 {\scriptsize [0.497, 0.647]} & 0.693 {\scriptsize [0.620, 0.761]} & 0.707 {\scriptsize [0.640, 0.774]} & 0.722 {\scriptsize [0.653, 0.790]} \\
    \bottomrule
  \end{tabular}
  }
\end{table*}

\begin{table*}[htbp]
  \centering
  \caption{Balanced Accuracy on the \textbf{\textsc{Resource}} task (ordering tests / referrals) across perturbation conditions and raters. Balanced Accuracy is computed as $(\text{TPR}+\text{TNR})/2$; 95\% bootstrap confidence intervals shown in brackets. Clinician values are pooled across all individual rater annotations.}
  \label{tab:balanced_accuracy_resource}
  \resizebox{\linewidth}{!}{%
  \setlength{\tabcolsep}{5pt}
  \begin{tabular}{lcccccc}
    \toprule
    \textbf{Perturbation} & \textbf{Clinician} & \texttt{GPT-4o} & \texttt{DeepSeek} & \texttt{MedGemma} & \texttt{Llama} & \texttt{Qwen} \\
    \midrule
    \rowcolor{gray!10}
    Baseline & \textbf{0.907} {\scriptsize [0.888, 0.924]} & \textbf{0.931} {\scriptsize [0.906, 0.956]} & \textbf{0.781} {\scriptsize [0.732, 0.827]} & \textbf{0.681} {\scriptsize [0.625, 0.732]} & \textbf{0.697} {\scriptsize [0.650, 0.744]} & \textbf{0.902} {\scriptsize [0.865, 0.936]} \\
    \midrule
    Gender Swap & 0.867 {\scriptsize [0.845, 0.888]} & 0.844 {\scriptsize [0.801, 0.883]} & 0.534 {\scriptsize [0.481, 0.586]} & 0.672 {\scriptsize [0.620, 0.723]} & 0.647 {\scriptsize [0.604, 0.692]} & 0.750 {\scriptsize [0.701, 0.798]} \\
    Gender Removal & 0.873 {\scriptsize [0.851, 0.894]} & 0.835 {\scriptsize [0.793, 0.878]} & 0.559 {\scriptsize [0.509, 0.610]} & 0.618 {\scriptsize [0.567, 0.668]} & 0.652 {\scriptsize [0.607, 0.694]} & 0.741 {\scriptsize [0.693, 0.788]} \\
    Colorful Tone & 0.814 {\scriptsize [0.781, 0.847]} & 0.734 {\scriptsize [0.666, 0.798]} & 0.569 {\scriptsize [0.499, 0.643]} & 0.669 {\scriptsize [0.606, 0.734]} & 0.698 {\scriptsize [0.638, 0.760]} & 0.715 {\scriptsize [0.647, 0.783]} \\
    Uncertain Tone & 0.871 {\scriptsize [0.844, 0.897]} & 0.805 {\scriptsize [0.741, 0.862]} & 0.589 {\scriptsize [0.519, 0.660]} & 0.698 {\scriptsize [0.634, 0.764]} & 0.687 {\scriptsize [0.628, 0.752]} & 0.707 {\scriptsize [0.633, 0.775]} \\
    \bottomrule
  \end{tabular}
  }
\end{table*}

% ─────────────────────────────────────────────────────────────
% Analysis paragraph
% ─────────────────────────────────────────────────────────────

Balanced Accuracy results (Tables~\ref{tab:balanced_accuracy_manage}--\ref{tab:balanced_accuracy_resource}) reinforce and sharpen the accuracy findings.
Clinicians maintain high Balanced Accuracy at baseline (0.837--0.925) and lose at most 0.093 under any perturbation, with gender rewrites costing 0.042 on average.
The LLMs separate into three regimes.
\texttt{GPT-4o} and \texttt{Qwen} start close to clinician level (0.830--0.953) but concede far more under perturbation, averaging 0.137 and 0.166 respectively under tone rewrites; \texttt{GPT-4o} takes the largest single tone hit, 0.217 on \textsc{Visit} under colorful language (0.953~$\to$~0.736).
\texttt{Llama} and \texttt{MedGemma} begin much lower (0.627--0.739) and have correspondingly less to lose: their average drop stays under 0.059 in every condition family, and \texttt{Llama} shows no mean change at all under tone ($-0.002$).
Their apparent robustness reflects a weak starting point rather than genuine stability.
\texttt{DeepSeek} degrades the most of any model, losing 0.194 on average under gender rewrites and 0.298 on \textsc{Visit} under gender removal (0.809~$\to$~0.511), a confidence interval that still includes chance.
Unlike the accuracy tables, tone is not uniformly the dominant failure mode: for \texttt{DeepSeek} and \texttt{Llama} gender rewrites cause larger drops in Balanced Accuracy than tone does, indicating that the two perturbation families stress different models in different ways.
No model--condition cell falls below chance in these results.

\subsection{Ablation Across Datasets}
\label{sec:ablation_datasets}
% =================================================================
% Dataset x perturbation performance section
% =================================================================

\begin{table*}[htbp]
  \centering
  \caption{Accuracy (\%) under the \textbf{Baseline} condition by dataset and task. Per-task values (\textsc{Manage}, \textsc{Visit}, \textsc{Resource}) and their macro mean (\textit{Avg}) are shown for each rater. The \textbf{Average} block aggregates across datasets. 95\% Wilson CIs in {\tiny brackets}; Average-block CIs over dataset means.}
  \label{tab:ds_baseline}
  \resizebox{\linewidth}{!}{%
  \setlength{\tabcolsep}{4pt}
  \begin{tabular}{llcccccc}
    \toprule
    \textbf{Dataset} & \textbf{Task} & \textbf{Clin.} & \texttt{GPT-4o} & \texttt{DeepSeek} & \texttt{MedGemma} & \texttt{Llama} & \texttt{Qwen} \\
    \midrule
    \multirow{4}{*}{\textbf{AskDocs}}
      & \textsc{Manage} & 84.8 {\tiny [81.4,87.7]} & 84.2 {\tiny [77.9,89.0]} & 60.1 {\tiny [52.5,67.3]} & 69.5 {\tiny [62.1,76.0]} & 88.5 {\tiny [82.7,92.5]} & 75.6 {\tiny [68.5,81.5]} \\
      & \textsc{Visit} & 90.9 {\tiny [88.0,93.1]} & 94.5 {\tiny [90.0,97.1]} & 81.0 {\tiny [74.3,86.3]} & 68.3 {\tiny [60.8,74.9]} & 85.5 {\tiny [79.3,90.0]} & 91.5 {\tiny [86.2,94.8]} \\
      & \textsc{Resource} & 88.7 {\tiny [85.6,91.2]} & 90.9 {\tiny [85.5,94.4]} & 78.5 {\tiny [71.6,84.1]} & 70.7 {\tiny [63.4,77.2]} & 84.2 {\tiny [77.9,89.0]} & 90.2 {\tiny [84.7,93.9]} \\
      & \textit{Avg} & \textit{88.1} {\tiny [84.6,91.6]} & \textit{89.9} {\tiny [84.0,95.8]} & \textit{73.2} {\tiny [60.3,86.1]} & \textit{69.5} {\tiny [68.1,70.9]} & \textit{86.1} {\tiny [83.6,88.5]} & \textit{85.8} {\tiny [75.8,95.8]} \\
    \cmidrule{1-8}
    \multirow{4}{*}{\textbf{OncQA}}
      & \textsc{Manage} & 82.0 {\tiny [75.7,87.0]} & 87.9 {\tiny [77.1,94.0]} & 81.0 {\tiny [69.1,89.1]} & 86.2 {\tiny [75.1,92.8]} & 89.7 {\tiny [79.2,95.2]} & 82.8 {\tiny [71.1,90.4]} \\
      & \textsc{Visit} & 95.5 {\tiny [91.4,97.7]} & 94.8 {\tiny [85.9,98.2]} & 84.5 {\tiny [73.1,91.6]} & 82.8 {\tiny [71.1,90.4]} & 91.4 {\tiny [81.4,96.3]} & 91.4 {\tiny [81.4,96.3]} \\
      & \textsc{Resource} & 92.1 {\tiny [87.2,95.3]} & 89.7 {\tiny [79.2,95.2]} & 84.5 {\tiny [73.1,91.6]} & 87.9 {\tiny [77.1,94.0]} & 93.1 {\tiny [83.6,97.3]} & 93.1 {\tiny [83.6,97.3]} \\
      & \textit{Avg} & \textit{89.9} {\tiny [81.9,97.8]} & \textit{90.8} {\tiny [86.7,94.9]} & \textit{83.3} {\tiny [81.1,85.6]} & \textit{85.6} {\tiny [82.7,88.6]} & \textit{91.4} {\tiny [89.4,93.3]} & \textit{89.1} {\tiny [82.8,95.4]} \\
    \cmidrule{1-8}
    \multirow{4}{*}{\textbf{SCT}}
      & \textsc{Manage} & 87.2 {\tiny [83.7,90.0]} & 84.5 {\tiny [77.8,89.4]} & 79.1 {\tiny [71.8,84.8]} & 87.8 {\tiny [81.6,92.2]} & 93.9 {\tiny [88.8,96.8]} & 94.6 {\tiny [89.7,97.2]} \\
      & \textsc{Visit} & 95.3 {\tiny [92.9,96.9]} & 96.6 {\tiny [92.3,98.5]} & 85.1 {\tiny [78.5,90.0]} & 68.9 {\tiny [61.1,75.8]} & 92.6 {\tiny [87.2,95.8]} & 96.6 {\tiny [92.3,98.5]} \\
      & \textsc{Resource} & 93.5 {\tiny [90.8,95.4]} & 95.3 {\tiny [90.6,97.7]} & 82.4 {\tiny [75.5,87.7]} & 74.3 {\tiny [66.7,80.7]} & 88.5 {\tiny [82.4,92.7]} & 95.3 {\tiny [90.6,97.7]} \\
      & \textit{Avg} & \textit{92.0} {\tiny [87.1,96.8]} & \textit{92.1} {\tiny [84.6,99.7]} & \textit{82.2} {\tiny [78.8,85.7]} & \textit{77.0} {\tiny [66.0,88.1]} & \textit{91.7} {\tiny [88.5,94.8]} & \textit{95.5} {\tiny [94.3,96.7]} \\
    \cmidrule{1-8}
    \multirow{4}{*}{\textbf{USMLE-Derm}}
      & \textsc{Manage} & 85.4 {\tiny [83.0,87.6]} & 85.7 {\tiny [81.2,89.2]} & 69.5 {\tiny [61.1,76.7]} & 73.2 {\tiny [65.2,79.9]} & 90.5 {\tiny [86.4,93.4]} & 84.5 {\tiny [78.7,88.9]} \\
      & \textsc{Visit} & 94.7 {\tiny [93.1,96.0]} & 97.7 {\tiny [95.3,98.9]} & 81.7 {\tiny [74.2,87.4]} & 67.9 {\tiny [59.5,75.3]} & 89.8 {\tiny [85.6,92.8]} & 91.2 {\tiny [86.3,94.4]} \\
      & \textsc{Resource} & 93.3 {\tiny [91.4,94.7]} & 94.7 {\tiny [91.5,96.7]} & 79.7 {\tiny [71.9,85.7]} & 73.2 {\tiny [64.9,80.2]} & 92.0 {\tiny [88.2,94.7]} & 92.2 {\tiny [87.6,95.2]} \\
      & \textit{Avg} & \textit{91.1} {\tiny [85.5,96.8]} & \textit{92.7} {\tiny [85.6,99.7]} & \textit{76.9} {\tiny [69.5,84.4]} & \textit{71.5} {\tiny [68.0,74.9]} & \textit{90.8} {\tiny [89.5,92.1]} & \textit{89.3} {\tiny [84.5,94.1]} \\
    \midrule
    \midrule
    \multirow{4}{*}{\textbf{Average}}
      & \textsc{Manage} & \textbf{84.9} {\tiny [82.8,86.9]} & \textbf{85.6} {\tiny [83.9,87.2]} & \textbf{72.4} {\tiny [63.0,81.9]} & \textbf{79.2} {\tiny [70.2,88.2]} & \textbf{90.6} {\tiny [88.3,92.9]} & \textbf{84.4} {\tiny [76.7,92.0]} \\
      & \textsc{Visit} & \textbf{94.1} {\tiny [92.0,96.2]} & \textbf{95.9} {\tiny [94.5,97.4]} & \textbf{83.1} {\tiny [81.1,85.1]} & \textbf{72.0} {\tiny [64.9,79.0]} & \textbf{89.8} {\tiny [86.7,92.8]} & \textbf{92.7} {\tiny [90.1,95.3]} \\
      & \textsc{Resource} & \textbf{91.9} {\tiny [89.7,94.1]} & \textbf{92.6} {\tiny [89.9,95.3]} & \textbf{81.3} {\tiny [78.6,83.9]} & \textbf{76.6} {\tiny [69.0,84.1]} & \textbf{89.5} {\tiny [85.6,93.4]} & \textbf{92.7} {\tiny [90.7,94.8]} \\
      & \textbf{\textit{Avg}} & \textbf{\textit{90.3}} {\tiny [87.7,92.9]} & \textbf{\textit{91.4}} {\tiny [88.6,94.1]} & \textbf{\textit{78.9}} {\tiny [74.8,83.0]} & \textbf{\textit{75.9}} {\tiny [71.4,80.4]} & \textbf{\textit{90.0}} {\tiny [88.3,91.6]} & \textbf{\textit{89.9}} {\tiny [86.5,93.3]} \\
    \bottomrule
  \end{tabular}
  }
\end{table*}

\begin{table*}[htbp]
  \centering
  \caption{Accuracy (\%) under the \textbf{Gender Swap} condition by dataset and task. Per-task values (\textsc{Manage}, \textsc{Visit}, \textsc{Resource}) and their macro mean (\textit{Avg}) are shown for each rater. The \textbf{Average} block aggregates across datasets. 95\% Wilson CIs in {\tiny brackets}; Average-block CIs over dataset means.}
  \label{tab:ds_gender_swap}
  \resizebox{\linewidth}{!}{%
  \setlength{\tabcolsep}{4pt}
  \begin{tabular}{llcccccc}
    \toprule
    \textbf{Dataset} & \textbf{Task} & \textbf{Clin.} & \texttt{GPT-4o} & \texttt{DeepSeek} & \texttt{MedGemma} & \texttt{Llama} & \texttt{Qwen} \\
    \midrule
    \multirow{4}{*}{\textbf{AskDocs}}
      & \textsc{Manage} & 77.9 {\tiny [74.1,81.4]} & 82.2 {\tiny [75.6,87.3]} & 62.2 {\tiny [54.6,69.3]} & 68.5 {\tiny [61.0,75.1]} & 81.2 {\tiny [74.6,86.4]} & 74.4 {\tiny [67.2,80.5]} \\
      & \textsc{Visit} & 88.5 {\tiny [85.3,91.0]} & 86.5 {\tiny [80.4,90.9]} & 51.8 {\tiny [44.2,59.3]} & 68.5 {\tiny [61.0,75.1]} & 77.0 {\tiny [70.0,82.7]} & 70.1 {\tiny [62.7,76.6]} \\
      & \textsc{Resource} & 84.8 {\tiny [81.4,87.7]} & 84.7 {\tiny [78.3,89.4]} & 59.8 {\tiny [52.1,67.0]} & 73.9 {\tiny [66.8,80.0]} & 76.4 {\tiny [69.3,82.2]} & 73.2 {\tiny [65.9,79.4]} \\
      & \textit{Avg} & \textit{83.7} {\tiny [77.7,89.8]} & \textit{84.5} {\tiny [82.0,86.9]} & \textit{57.9} {\tiny [51.8,64.1]} & \textit{70.3} {\tiny [66.7,73.9]} & \textit{78.2} {\tiny [75.2,81.2]} & \textit{72.6} {\tiny [70.1,75.0]} \\
    \cmidrule{1-8}
    \multirow{4}{*}{\textbf{OncQA}}
      & \textsc{Manage} & 74.3 {\tiny [67.3,80.2]} & 84.2 {\tiny [72.6,91.5]} & 77.6 {\tiny [65.3,86.4]} & 86.2 {\tiny [75.1,92.8]} & 81.0 {\tiny [69.1,89.1]} & 81.0 {\tiny [69.1,89.1]} \\
      & \textsc{Visit} & 94.3 {\tiny [89.8,96.9]} & 89.5 {\tiny [78.9,95.1]} & 77.6 {\tiny [65.3,86.4]} & 82.8 {\tiny [71.1,90.4]} & 89.7 {\tiny [79.2,95.2]} & 81.0 {\tiny [69.1,89.1]} \\
      & \textsc{Resource} & 93.1 {\tiny [88.4,96.0]} & 84.2 {\tiny [72.6,91.5]} & 72.4 {\tiny [59.8,82.2]} & 89.7 {\tiny [79.2,95.2]} & 89.7 {\tiny [79.2,95.2]} & 79.3 {\tiny [67.2,87.7]} \\
      & \textit{Avg} & \textit{87.2} {\tiny [74.5,99.9]} & \textit{86.0} {\tiny [82.5,89.4]} & \textit{75.9} {\tiny [72.5,79.2]} & \textit{86.2} {\tiny [82.3,90.1]} & \textit{86.8} {\tiny [81.1,92.4]} & \textit{80.5} {\tiny [79.3,81.6]} \\
    \cmidrule{1-8}
    \multirow{4}{*}{\textbf{SCT}}
      & \textsc{Manage} & 86.8 {\tiny [83.3,89.6]} & 92.5 {\tiny [87.0,95.7]} & 78.1 {\tiny [70.7,84.0]} & 87.7 {\tiny [81.4,92.1]} & 91.1 {\tiny [85.4,94.7]} & 92.5 {\tiny [87.0,95.7]} \\
      & \textsc{Visit} & 92.7 {\tiny [89.9,94.8]} & 89.0 {\tiny [82.9,93.1]} & 58.2 {\tiny [50.1,65.9]} & 65.8 {\tiny [57.7,73.0]} & 91.8 {\tiny [86.2,95.2]} & 89.7 {\tiny [83.7,93.7]} \\
      & \textsc{Resource} & 89.0 {\tiny [85.8,91.6]} & 91.1 {\tiny [85.4,94.7]} & 65.8 {\tiny [57.7,73.0]} & 69.2 {\tiny [61.3,76.1]} & 89.7 {\tiny [83.7,93.7]} & 88.4 {\tiny [82.1,92.6]} \\
      & \textit{Avg} & \textit{89.5} {\tiny [86.1,92.9]} & \textit{90.9} {\tiny [88.9,92.8]} & \textit{67.4} {\tiny [56.0,78.7]} & \textit{74.2} {\tiny [60.9,87.5]} & \textit{90.9} {\tiny [89.7,92.1]} & \textit{90.2} {\tiny [87.8,92.6]} \\
    \cmidrule{1-8}
    \multirow{4}{*}{\textbf{USMLE-Derm}}
      & \textsc{Manage} & 85.6 {\tiny [83.2,87.8]} & 88.0 {\tiny [83.8,91.2]} & 67.0 {\tiny [60.0,73.4]} & 63.6 {\tiny [56.5,70.2]} & 89.6 {\tiny [85.7,92.6]} & 79.0 {\tiny [73.3,83.8]} \\
      & \textsc{Visit} & 92.5 {\tiny [90.6,94.1]} & 91.7 {\tiny [88.0,94.3]} & 62.0 {\tiny [54.8,68.7]} & 70.1 {\tiny [63.1,76.2]} & 90.0 {\tiny [86.0,92.9]} & 78.3 {\tiny [72.5,83.1]} \\
      & \textsc{Resource} & 92.5 {\tiny [90.6,94.1]} & 92.0 {\tiny [88.4,94.6]} & 56.6 {\tiny [49.3,63.6]} & 67.2 {\tiny [60.1,73.6]} & 91.6 {\tiny [87.9,94.3]} & 81.7 {\tiny [76.1,86.1]} \\
      & \textit{Avg} & \textit{90.2} {\tiny [85.7,94.7]} & \textit{90.6} {\tiny [88.0,93.1]} & \textit{61.9} {\tiny [56.0,67.8]} & \textit{67.0} {\tiny [63.3,70.6]} & \textit{90.4} {\tiny [89.2,91.6]} & \textit{79.7} {\tiny [77.6,81.7]} \\
    \midrule
    \midrule
    \multirow{4}{*}{\textbf{Average}}
      & \textsc{Manage} & \textbf{81.2} {\tiny [75.2,87.1]} & \textbf{86.7} {\tiny [82.3,91.2]} & \textbf{71.2} {\tiny [63.5,79.0]} & \textbf{76.5} {\tiny [64.5,88.5]} & \textbf{85.7} {\tiny [80.5,91.0]} & \textbf{81.7} {\tiny [74.2,89.3]} \\
      & \textsc{Visit} & \textbf{92.0} {\tiny [89.6,94.4]} & \textbf{89.2} {\tiny [87.1,91.2]} & \textbf{62.4} {\tiny [51.7,73.1]} & \textbf{71.8} {\tiny [64.4,79.2]} & \textbf{87.1} {\tiny [80.4,93.8]} & \textbf{79.8} {\tiny [71.9,87.7]} \\
      & \textsc{Resource} & \textbf{89.9} {\tiny [86.1,93.6]} & \textbf{88.0} {\tiny [83.9,92.0]} & \textbf{63.6} {\tiny [56.8,70.5]} & \textbf{75.0} {\tiny [65.0,85.0]} & \textbf{86.8} {\tiny [79.9,93.8]} & \textbf{80.6} {\tiny [74.5,86.8]} \\
      & \textbf{\textit{Avg}} & \textbf{\textit{87.7}} {\tiny [84.1,91.2]} & \textbf{\textit{88.0}} {\tiny [86.0,90.0]} & \textbf{\textit{65.7}} {\tiny [60.7,70.8]} & \textbf{\textit{74.4}} {\tiny [69.1,79.8]} & \textbf{\textit{86.6}} {\tiny [83.2,89.9]} & \textbf{\textit{80.7}} {\tiny [76.9,84.5]} \\
    \bottomrule
  \end{tabular}
  }
\end{table*}

\begin{table*}[htbp]
  \centering
  \caption{Accuracy (\%) under the \textbf{Gender Removal} condition by dataset and task. Per-task values (\textsc{Manage}, \textsc{Visit}, \textsc{Resource}) and their macro mean (\textit{Avg}) are shown for each rater. The \textbf{Average} block aggregates across datasets. 95\% Wilson CIs in {\tiny brackets}; Average-block CIs over dataset means.}
  \label{tab:ds_gender_removal}
  \resizebox{\linewidth}{!}{%
  \setlength{\tabcolsep}{4pt}
  \begin{tabular}{llcccccc}
    \toprule
    \textbf{Dataset} & \textbf{Task} & \textbf{Clin.} & \texttt{GPT-4o} & \texttt{DeepSeek} & \texttt{MedGemma} & \texttt{Llama} & \texttt{Qwen} \\
    \midrule
    \multirow{4}{*}{\textbf{AskDocs}}
      & \textsc{Manage} & 80.6 {\tiny [76.9,83.8]} & 81.8 {\tiny [75.2,87.0]} & 49.7 {\tiny [42.2,57.2]} & 63.0 {\tiny [55.4,70.0]} & 82.4 {\tiny [75.9,87.5]} & 74.5 {\tiny [67.4,80.6]} \\
      & \textsc{Visit} & 86.5 {\tiny [83.2,89.2]} & 82.4 {\tiny [75.9,87.5]} & 56.4 {\tiny [48.7,63.7]} & 70.3 {\tiny [62.9,76.7]} & 81.8 {\tiny [75.2,87.0]} & 71.5 {\tiny [64.2,77.9]} \\
      & \textsc{Resource} & 83.4 {\tiny [79.9,86.4]} & 84.2 {\tiny [77.9,89.0]} & 60.0 {\tiny [52.4,67.2]} & 69.7 {\tiny [62.3,76.2]} & 77.0 {\tiny [70.0,82.7]} & 73.9 {\tiny [66.8,80.0]} \\
      & \textit{Avg} & \textit{83.5} {\tiny [80.2,86.8]} & \textit{82.8} {\tiny [81.4,84.3]} & \textit{55.4} {\tiny [49.4,61.3]} & \textit{67.7} {\tiny [63.1,72.2]} & \textit{80.4} {\tiny [77.0,83.8]} & \textit{73.3} {\tiny [71.5,75.1]} \\
    \cmidrule{1-8}
    \multirow{4}{*}{\textbf{OncQA}}
      & \textsc{Manage} & 73.0 {\tiny [65.9,79.0]} & 70.7 {\tiny [58.0,80.8]} & 81.0 {\tiny [69.1,89.1]} & 87.9 {\tiny [77.1,94.0]} & 87.7 {\tiny [76.8,93.9]} & 77.6 {\tiny [65.3,86.4]} \\
      & \textsc{Visit} & 90.2 {\tiny [84.9,93.8]} & 84.5 {\tiny [73.1,91.6]} & 75.9 {\tiny [63.5,85.0]} & 81.0 {\tiny [69.1,89.1]} & 93.0 {\tiny [83.3,97.2]} & 81.0 {\tiny [69.1,89.1]} \\
      & \textsc{Resource} & 90.2 {\tiny [84.9,93.8]} & 79.3 {\tiny [67.2,87.7]} & 65.5 {\tiny [52.7,76.4]} & 84.5 {\tiny [73.1,91.6]} & 93.0 {\tiny [83.3,97.2]} & 84.5 {\tiny [73.1,91.6]} \\
      & \textit{Avg} & \textit{84.5} {\tiny [73.2,95.7]} & \textit{78.2} {\tiny [70.3,86.0]} & \textit{74.1} {\tiny [65.2,83.1]} & \textit{84.5} {\tiny [80.6,88.4]} & \textit{91.2} {\tiny [87.8,94.7]} & \textit{81.0} {\tiny [77.1,84.9]} \\
    \cmidrule{1-8}
    \multirow{4}{*}{\textbf{SCT}}
      & \textsc{Manage} & 87.8 {\tiny [84.5,90.6]} & 95.3 {\tiny [90.6,97.7]} & 74.3 {\tiny [66.7,80.7]} & 88.5 {\tiny [82.4,92.7]} & 91.9 {\tiny [86.4,95.3]} & 92.6 {\tiny [87.2,95.8]} \\
      & \textsc{Visit} & 94.1 {\tiny [91.6,96.0]} & 89.2 {\tiny [83.2,93.2]} & 57.4 {\tiny [49.4,65.1]} & 59.5 {\tiny [51.4,67.0]} & 92.6 {\tiny [87.2,95.8]} & 87.8 {\tiny [81.6,92.2]} \\
      & \textsc{Resource} & 92.3 {\tiny [89.5,94.5]} & 91.9 {\tiny [86.4,95.3]} & 63.5 {\tiny [55.5,70.8]} & 66.2 {\tiny [58.3,73.3]} & 87.8 {\tiny [81.6,92.2]} & 89.2 {\tiny [83.2,93.2]} \\
      & \textit{Avg} & \textit{91.4} {\tiny [87.8,95.1]} & \textit{92.1} {\tiny [88.7,95.6]} & \textit{65.1} {\tiny [55.4,74.8]} & \textit{71.4} {\tiny [54.2,88.6]} & \textit{90.8} {\tiny [87.9,93.7]} & \textit{89.9} {\tiny [87.1,92.6]} \\
    \cmidrule{1-8}
    \multirow{4}{*}{\textbf{USMLE-Derm}}
      & \textsc{Manage} & 84.9 {\tiny [82.4,87.1]} & 88.3 {\tiny [84.2,91.5]} & 65.3 {\tiny [59.7,70.5]} & 68.1 {\tiny [62.6,73.2]} & 89.6 {\tiny [85.7,92.6]} & 76.4 {\tiny [71.2,80.8]} \\
      & \textsc{Visit} & 93.4 {\tiny [91.6,94.8]} & 91.3 {\tiny [87.6,94.0]} & 59.2 {\tiny [53.5,64.6]} & 63.6 {\tiny [58.0,68.9]} & 89.6 {\tiny [85.7,92.6]} & 73.3 {\tiny [68.0,78.0]} \\
      & \textsc{Resource} & 92.2 {\tiny [90.2,93.8]} & 91.0 {\tiny [87.2,93.7]} & 63.3 {\tiny [57.6,68.6]} & 69.4 {\tiny [63.9,74.4]} & 92.0 {\tiny [88.3,94.5]} & 76.7 {\tiny [71.5,81.1]} \\
      & \textit{Avg} & \textit{90.2} {\tiny [85.0,95.4]} & \textit{90.2} {\tiny [88.4,92.1]} & \textit{62.6} {\tiny [59.1,66.1]} & \textit{67.0} {\tiny [63.6,70.5]} & \textit{90.4} {\tiny [88.9,91.9]} & \textit{75.5} {\tiny [73.3,77.6]} \\
    \midrule
    \midrule
    \multirow{4}{*}{\textbf{Average}}
      & \textsc{Manage} & \textbf{81.6} {\tiny [75.3,87.9]} & \textbf{84.0} {\tiny [73.8,94.3]} & \textbf{67.6} {\tiny [54.3,80.9]} & \textbf{76.9} {\tiny [63.9,89.9]} & \textbf{87.9} {\tiny [84.0,91.9]} & \textbf{80.3} {\tiny [72.1,88.4]} \\
      & \textsc{Visit} & \textbf{91.1} {\tiny [87.6,94.5]} & \textbf{86.9} {\tiny [82.8,90.9]} & \textbf{62.2} {\tiny [53.2,71.2]} & \textbf{68.6} {\tiny [59.4,77.8]} & \textbf{89.3} {\tiny [84.2,94.3]} & \textbf{78.4} {\tiny [71.1,85.8]} \\
      & \textsc{Resource} & \textbf{89.5} {\tiny [85.4,93.6]} & \textbf{86.6} {\tiny [80.8,92.4]} & \textbf{63.1} {\tiny [60.8,65.3]} & \textbf{72.4} {\tiny [64.4,80.5]} & \textbf{87.4} {\tiny [80.3,94.6]} & \textbf{81.1} {\tiny [74.2,87.9]} \\
      & \textbf{\textit{Avg}} & \textbf{\textit{87.4}} {\tiny [83.9,90.9]} & \textbf{\textit{85.8}} {\tiny [82.0,89.7]} & \textbf{\textit{64.3}} {\tiny [59.2,69.4]} & \textbf{\textit{72.6}} {\tiny [66.9,78.4]} & \textbf{\textit{88.2}} {\tiny [85.3,91.1]} & \textbf{\textit{79.9}} {\tiny [76.0,83.9]} \\
    \bottomrule
  \end{tabular}
  }
\end{table*}

\begin{table*}[htbp]
  \centering
  \caption{Accuracy (\%) under the \textbf{Colorful Tone} condition by dataset and task. Per-task values (\textsc{Manage}, \textsc{Visit}, \textsc{Resource}) and their macro mean (\textit{Avg}) are shown for each rater. The \textbf{Average} block aggregates across datasets. 95\% Wilson CIs in {\tiny brackets}; Average-block CIs over dataset means. Only \textbf{AskDocs} and \textbf{OncQA} include tone-perturbed annotations.}
  \label{tab:ds_colorful}
  \resizebox{\linewidth}{!}{%
  \setlength{\tabcolsep}{4pt}
  \begin{tabular}{llcccccc}
    \toprule
    \textbf{Dataset} & \textbf{Task} & \textbf{Clin.} & \texttt{GPT-4o} & \texttt{DeepSeek} & \texttt{MedGemma} & \texttt{Llama} & \texttt{Qwen} \\
    \midrule
    \multirow{4}{*}{\textbf{AskDocs}}
      & \textsc{Manage} & 80.2 {\tiny [76.4,83.5]} & 81.6 {\tiny [74.9,86.8]} & 60.4 {\tiny [52.7,67.5]} & 58.5 {\tiny [50.9,65.8]} & 82.4 {\tiny [75.9,87.5]} & 68.1 {\tiny [60.6,74.8]} \\
      & \textsc{Visit} & 85.7 {\tiny [82.3,88.5]} & 78.5 {\tiny [71.6,84.1]} & 62.0 {\tiny [54.3,69.1]} & 65.9 {\tiny [58.3,72.7]} & 81.8 {\tiny [75.2,87.0]} & 74.8 {\tiny [67.7,80.9]} \\
      & \textsc{Resource} & 83.5 {\tiny [79.9,86.5]} & 79.1 {\tiny [72.3,84.7]} & 61.3 {\tiny [53.7,68.5]} & 68.3 {\tiny [60.8,74.9]} & 78.2 {\tiny [71.3,83.8]} & 71.2 {\tiny [63.8,77.6]} \\
      & \textit{Avg} & \textit{83.1} {\tiny [80.0,86.3]} & \textit{79.8} {\tiny [77.9,81.6]} & \textit{61.2} {\tiny [60.3,62.1]} & \textit{64.2} {\tiny [58.5,70.0]} & \textit{80.8} {\tiny [78.2,83.4]} & \textit{71.4} {\tiny [67.5,75.2]} \\
    \cmidrule{1-8}
    \multirow{4}{*}{\textbf{OncQA}}
      & \textsc{Manage} & 73.9 {\tiny [66.9,79.8]} & 77.6 {\tiny [65.3,86.4]} & 78.9 {\tiny [66.7,87.5]} & 86.2 {\tiny [75.1,92.8]} & 81.5 {\tiny [69.2,89.6]} & 58.6 {\tiny [45.8,70.4]} \\
      & \textsc{Visit} & 93.8 {\tiny [89.2,96.5]} & 86.2 {\tiny [75.1,92.8]} & 61.4 {\tiny [48.4,72.9]} & 86.0 {\tiny [74.7,92.7]} & 88.9 {\tiny [77.8,94.8]} & 74.1 {\tiny [61.6,83.7]} \\
      & \textsc{Resource} & 87.5 {\tiny [81.8,91.6]} & 77.6 {\tiny [65.3,86.4]} & 61.4 {\tiny [48.4,72.9]} & 89.5 {\tiny [78.9,95.1]} & 90.7 {\tiny [80.1,96.0]} & 75.9 {\tiny [63.5,85.0]} \\
      & \textit{Avg} & \textit{85.0} {\tiny [73.5,96.5]} & \textit{80.5} {\tiny [74.8,86.1]} & \textit{67.3} {\tiny [55.8,78.7]} & \textit{87.2} {\tiny [85.0,89.4]} & \textit{87.0} {\tiny [81.5,92.6]} & \textit{69.5} {\tiny [58.8,80.3]} \\
    \midrule
    \midrule
    \multirow{4}{*}{\textbf{Average}}
      & \textsc{Manage} & \textbf{77.0} {\tiny [70.8,83.2]} & \textbf{79.6} {\tiny [75.7,83.5]} & \textbf{69.7} {\tiny [51.4,87.9]} & \textbf{72.4} {\tiny [45.3,99.5]} & \textbf{82.0} {\tiny [81.0,82.9]} & \textbf{63.4} {\tiny [54.1,72.6]} \\
      & \textsc{Visit} & \textbf{89.7} {\tiny [81.9,97.6]} & \textbf{82.4} {\tiny [74.8,89.9]} & \textbf{61.7} {\tiny [61.1,62.2]} & \textbf{75.9} {\tiny [56.2,95.6]} & \textbf{85.4} {\tiny [78.4,92.3]} & \textbf{74.5} {\tiny [73.8,75.2]} \\
      & \textsc{Resource} & \textbf{85.5} {\tiny [81.5,89.4]} & \textbf{78.4} {\tiny [76.8,79.9]} & \textbf{61.4} {\tiny [61.3,61.4]} & \textbf{78.9} {\tiny [58.1,99.6]} & \textbf{84.5} {\tiny [72.2,96.8]} & \textbf{73.5} {\tiny [68.9,78.1]} \\
      & \textbf{\textit{Avg}} & \textbf{\textit{84.1}} {\tiny [78.7,89.5]} & \textbf{\textit{80.1}} {\tiny [77.4,82.8]} & \textbf{\textit{64.2}} {\tiny [58.5,70.0]} & \textbf{\textit{75.7}} {\tiny [65.3,86.2]} & \textbf{\textit{83.9}} {\tiny [80.1,87.8]} & \textbf{\textit{70.5}} {\tiny [65.3,75.6]} \\
    \bottomrule
  \end{tabular}
  }
\end{table*}

\begin{table*}[htbp]
  \centering
  \caption{Accuracy (\%) under the \textbf{Uncertain Tone} condition by dataset and task. Per-task values (\textsc{Manage}, \textsc{Visit}, \textsc{Resource}) and their macro mean (\textit{Avg}) are shown for each rater. The \textbf{Average} block aggregates across datasets. 95\% Wilson CIs in {\tiny brackets}; Average-block CIs over dataset means. Only \textbf{AskDocs} and \textbf{OncQA} include tone-perturbed annotations.}
  \label{tab:ds_uncertain}
  \resizebox{\linewidth}{!}{%
  \setlength{\tabcolsep}{4pt}
  \begin{tabular}{llcccccc}
    \toprule
    \textbf{Dataset} & \textbf{Task} & \textbf{Clin.} & \texttt{GPT-4o} & \texttt{DeepSeek} & \texttt{MedGemma} & \texttt{Llama} & \texttt{Qwen} \\
    \midrule
    \multirow{4}{*}{\textbf{AskDocs}}
      & \textsc{Manage} & 81.0 {\tiny [77.3,84.2]} & 83.6 {\tiny [77.2,88.5]} & 65.5 {\tiny [57.9,72.3]} & 68.5 {\tiny [61.0,75.1]} & 80.0 {\tiny [73.2,85.4]} & 73.3 {\tiny [66.1,79.5]} \\
      & \textsc{Visit} & 89.7 {\tiny [86.7,92.1]} & 81.2 {\tiny [74.6,86.4]} & 58.2 {\tiny [50.6,65.4]} & 64.8 {\tiny [57.3,71.7]} & 79.4 {\tiny [72.6,84.9]} & 72.1 {\tiny [64.8,78.4]} \\
      & \textsc{Resource} & 84.8 {\tiny [81.4,87.7]} & 80.0 {\tiny [73.2,85.4]} & 60.6 {\tiny [53.0,67.7]} & 70.3 {\tiny [62.9,76.7]} & 77.0 {\tiny [70.0,82.7]} & 72.1 {\tiny [64.8,78.4]} \\
      & \textit{Avg} & \textit{85.2} {\tiny [80.3,90.1]} & \textit{81.6} {\tiny [79.5,83.7]} & \textit{61.4} {\tiny [57.2,65.6]} & \textit{67.9} {\tiny [64.7,71.0]} & \textit{78.8} {\tiny [77.0,80.6]} & \textit{72.5} {\tiny [71.7,73.3]} \\
    \cmidrule{1-8}
    \multirow{4}{*}{\textbf{OncQA}}
      & \textsc{Manage} & 80.6 {\tiny [74.1,85.8]} & 75.9 {\tiny [63.5,85.0]} & 75.4 {\tiny [62.9,84.8]} & 77.2 {\tiny [64.8,86.2]} & 81.5 {\tiny [69.2,89.6]} & 63.2 {\tiny [50.2,74.5]} \\
      & \textsc{Visit} & 93.7 {\tiny [89.1,96.5]} & 82.8 {\tiny [71.1,90.4]} & 61.4 {\tiny [48.4,72.9]} & 94.7 {\tiny [85.6,98.2]} & 90.7 {\tiny [80.1,96.0]} & 68.4 {\tiny [55.5,79.0]} \\
      & \textsc{Resource} & 91.4 {\tiny [86.3,94.7]} & 86.2 {\tiny [75.1,92.8]} & 59.6 {\tiny [46.7,71.4]} & 94.7 {\tiny [85.6,98.2]} & 90.7 {\tiny [80.1,96.0]} & 78.9 {\tiny [66.7,87.5]} \\
      & \textit{Avg} & \textit{88.6} {\tiny [80.6,96.5]} & \textit{81.6} {\tiny [75.6,87.6]} & \textit{65.5} {\tiny [55.7,75.3]} & \textit{88.9} {\tiny [77.4,100.0]} & \textit{87.7} {\tiny [81.6,93.7]} & \textit{70.2} {\tiny [61.1,79.3]} \\
    \midrule
    \midrule
    \multirow{4}{*}{\textbf{Average}}
      & \textsc{Manage} & \textbf{80.8} {\tiny [80.4,81.2]} & \textbf{79.7} {\tiny [72.1,87.4]} & \textbf{70.4} {\tiny [60.7,80.2]} & \textbf{72.8} {\tiny [64.3,81.4]} & \textbf{80.7} {\tiny [79.3,82.2]} & \textbf{68.2} {\tiny [58.3,78.2]} \\
      & \textsc{Visit} & \textbf{91.7} {\tiny [87.8,95.6]} & \textbf{82.0} {\tiny [80.5,83.5]} & \textbf{59.8} {\tiny [56.6,62.9]} & \textbf{79.8} {\tiny [50.5,100.0]} & \textbf{85.1} {\tiny [73.9,96.2]} & \textbf{70.3} {\tiny [66.6,73.9]} \\
      & \textsc{Resource} & \textbf{88.1} {\tiny [81.7,94.6]} & \textbf{83.1} {\tiny [77.0,89.2]} & \textbf{60.1} {\tiny [59.2,61.1]} & \textbf{82.5} {\tiny [58.6,100.0]} & \textbf{83.9} {\tiny [70.4,97.4]} & \textbf{75.5} {\tiny [68.8,82.2]} \\
      & \textbf{\textit{Avg}} & \textbf{\textit{86.9}} {\tiny [82.4,91.3]} & \textbf{\textit{81.6}} {\tiny [78.8,84.4]} & \textbf{\textit{63.5}} {\tiny [58.4,68.5]} & \textbf{\textit{78.4}} {\tiny [67.8,89.0]} & \textbf{\textit{83.2}} {\tiny [78.4,88.0]} & \textbf{\textit{71.4}} {\tiny [67.1,75.6]} \\
    \bottomrule
  \end{tabular}
  }
\end{table*}

% -----------------------------------------------------------------
% Analysis paragraphs
% -----------------------------------------------------------------

\paragraph{Baseline heterogeneity across datasets.}
Table~\ref{tab:ds_baseline} establishes the reference performance landscape and reveals substantial variation across the four benchmark dataset groups. \textbf{AskDocs} is the most difficult for every rater: clinician accuracy averages 88.1\% and LLM averages range from 69.5\% (\texttt{MedGemma}) to 89.9\% (\texttt{GPT-4o}), reflecting the informal, patient-authored register of real forum posts. \textbf{SCT} and \textbf{USMLE-Derm} produce the highest clinician scores (92.0\% and 91.1\%), likely because their standardized vignette format aligns well with expert consensus. \textbf{OncQA} shows a clear task asymmetry: clinicians reach 95.5\% on \textsc{Visit} and 92.1\% on \textsc{Resource}, indicating strong consensus on oncology referral decisions, while \textsc{Manage} lags at 82.0\%, where the self-care versus intervention boundary is more contested. Across datasets, \texttt{MedGemma} and \texttt{DeepSeek} trail the other LLMs (macro averages 75.9\% and 78.9\%), while \texttt{GPT-4o} (91.4\%) matches the clinician panel (90.3\%).

\paragraph{Perturbation effects and cross-dataset heterogeneity.}
Tables~\ref{tab:ds_gender_swap}--\ref{tab:ds_uncertain} show that the magnitude of perturbation-induced degradation varies substantially by dataset. For gender perturbations, \textbf{AskDocs} shows the largest drops for most models: \texttt{GPT-4o} falls 2.0 pp on \textsc{Manage} under gender swap in AskDocs while gaining 8.0 pp on the same task in SCT, suggesting that the less-anchored register of patient posts amplifies sensitivity to identity cues. \texttt{DeepSeek} is the exception, losing 7.5--15.3 pp under gender swap depending on the corpus and more than 14 pp in 3 of the 4, including the standardized ones, so its fragility is not confined to the patient-facing corpora. Under gender swap, \texttt{MedGemma} changes only marginally on the two patient-facing corpora, gaining slightly in both. Tone perturbations are available only for AskDocs and OncQA, and their contrast is instructive: OncQA sustains higher clinician accuracy under both tone conditions (85.0\% and above), and \texttt{MedGemma} is actually more accurate there under tone than at baseline. \texttt{Qwen} shows the most dataset-sensitive degradation under tone: it loses 19.6 pp on OncQA against 14.4 pp on AskDocs under colorful language, falling to 58.6\% on OncQA \textsc{Manage} from an 82.8\% baseline. These patterns indicate that perturbation robustness is not a fixed model property but interacts with the difficulty and linguistic domain of the underlying dataset.

\begin{table*}[htbp]
  \centering
  \caption{Accuracy (\%) under the \textbf{Baseline} condition by dataset and task. Per-task values (\textsc{Manage}, \textsc{Visit}, \textsc{Resource}) and their macro mean (\textit{Avg}) are shown for each rater. The \textbf{Average} block aggregates across datasets. 95\% Wilson CIs in {\tiny brackets}; Average-block CIs over dataset means.}
  \label{tab:ds_baseline}
  \resizebox{\linewidth}{!}{%
  \setlength{\tabcolsep}{4pt}
  \begin{tabular}{llcccccc}
    \toprule
    \textbf{Dataset} & \textbf{Task} & \textbf{Clin.} & \texttt{GPT-4o} & \texttt{DeepSeek} & \texttt{MedGemma} & \texttt{Llama} & \texttt{Qwen} \\
    \midrule
    \multirow{4}{*}{\textbf{AskDocs}}
      & \textsc{Manage} & 84.8 {\tiny [81.4,87.7]} & 84.2 {\tiny [77.9,89.0]} & 60.1 {\tiny [52.5,67.3]} & 69.5 {\tiny [62.1,76.0]} & 88.5 {\tiny [82.7,92.5]} & 75.6 {\tiny [68.5,81.5]} \\
      & \textsc{Visit} & 90.9 {\tiny [88.0,93.1]} & 94.5 {\tiny [90.0,97.1]} & 81.0 {\tiny [74.3,86.3]} & 68.3 {\tiny [60.8,74.9]} & 85.5 {\tiny [79.3,90.0]} & 91.5 {\tiny [86.2,94.8]} \\
      & \textsc{Resource} & 88.7 {\tiny [85.6,91.2]} & 90.9 {\tiny [85.5,94.4]} & 78.5 {\tiny [71.6,84.1]} & 70.7 {\tiny [63.4,77.2]} & 84.2 {\tiny [77.9,89.0]} & 90.2 {\tiny [84.7,93.9]} \\
      & \textit{Avg} & \textit{88.1} {\tiny [84.6,91.6]} & \textit{89.9} {\tiny [84.0,95.8]} & \textit{73.2} {\tiny [60.3,86.1]} & \textit{69.5} {\tiny [68.1,70.9]} & \textit{86.1} {\tiny [83.6,88.5]} & \textit{85.8} {\tiny [75.8,95.8]} \\
    \cmidrule{1-8}
    \multirow{4}{*}{\textbf{OncQA}}
      & \textsc{Manage} & 82.0 {\tiny [75.7,87.0]} & 87.9 {\tiny [77.1,94.0]} & 81.0 {\tiny [69.1,89.1]} & 86.2 {\tiny [75.1,92.8]} & 89.7 {\tiny [79.2,95.2]} & 82.8 {\tiny [71.1,90.4]} \\
      & \textsc{Visit} & 95.5 {\tiny [91.4,97.7]} & 94.8 {\tiny [85.9,98.2]} & 84.5 {\tiny [73.1,91.6]} & 82.8 {\tiny [71.1,90.4]} & 91.4 {\tiny [81.4,96.3]} & 91.4 {\tiny [81.4,96.3]} \\
      & \textsc{Resource} & 92.1 {\tiny [87.2,95.3]} & 89.7 {\tiny [79.2,95.2]} & 84.5 {\tiny [73.1,91.6]} & 87.9 {\tiny [77.1,94.0]} & 93.1 {\tiny [83.6,97.3]} & 93.1 {\tiny [83.6,97.3]} \\
      & \textit{Avg} & \textit{89.9} {\tiny [81.9,97.8]} & \textit{90.8} {\tiny [86.7,94.9]} & \textit{83.3} {\tiny [81.1,85.6]} & \textit{85.6} {\tiny [82.7,88.6]} & \textit{91.4} {\tiny [89.4,93.3]} & \textit{89.1} {\tiny [82.8,95.4]} \\
    \cmidrule{1-8}
    \multirow{4}{*}{\textbf{SCT}}
      & \textsc{Manage} & 87.2 {\tiny [83.7,90.0]} & 84.5 {\tiny [77.8,89.4]} & 79.1 {\tiny [71.8,84.8]} & 87.8 {\tiny [81.6,92.2]} & 93.9 {\tiny [88.8,96.8]} & 94.6 {\tiny [89.7,97.2]} \\
      & \textsc{Visit} & 95.3 {\tiny [92.9,96.9]} & 96.6 {\tiny [92.3,98.5]} & 85.1 {\tiny [78.5,90.0]} & 68.9 {\tiny [61.1,75.8]} & 92.6 {\tiny [87.2,95.8]} & 96.6 {\tiny [92.3,98.5]} \\
      & \textsc{Resource} & 93.5 {\tiny [90.8,95.4]} & 95.3 {\tiny [90.6,97.7]} & 82.4 {\tiny [75.5,87.7]} & 74.3 {\tiny [66.7,80.7]} & 88.5 {\tiny [82.4,92.7]} & 95.3 {\tiny [90.6,97.7]} \\
      & \textit{Avg} & \textit{92.0} {\tiny [87.1,96.8]} & \textit{92.1} {\tiny [84.6,99.7]} & \textit{82.2} {\tiny [78.8,85.7]} & \textit{77.0} {\tiny [66.0,88.1]} & \textit{91.7} {\tiny [88.5,94.8]} & \textit{95.5} {\tiny [94.3,96.7]} \\
    \cmidrule{1-8}
    \multirow{4}{*}{\textbf{USMLE-Derm}}
      & \textsc{Manage} & 85.4 {\tiny [83.0,87.6]} & 85.7 {\tiny [81.2,89.2]} & 69.5 {\tiny [61.1,76.7]} & 73.2 {\tiny [65.2,79.9]} & 90.5 {\tiny [86.4,93.4]} & 84.5 {\tiny [78.7,88.9]} \\
      & \textsc{Visit} & 94.7 {\tiny [93.1,96.0]} & 97.7 {\tiny [95.3,98.9]} & 81.7 {\tiny [74.2,87.4]} & 67.9 {\tiny [59.5,75.3]} & 89.8 {\tiny [85.6,92.8]} & 91.2 {\tiny [86.3,94.4]} \\
      & \textsc{Resource} & 93.3 {\tiny [91.4,94.7]} & 94.7 {\tiny [91.5,96.7]} & 79.7 {\tiny [71.9,85.7]} & 73.2 {\tiny [64.9,80.2]} & 92.0 {\tiny [88.2,94.7]} & 92.2 {\tiny [87.6,95.2]} \\
      & \textit{Avg} & \textit{91.1} {\tiny [85.5,96.8]} & \textit{92.7} {\tiny [85.6,99.7]} & \textit{76.9} {\tiny [69.5,84.4]} & \textit{71.5} {\tiny [68.0,74.9]} & \textit{90.8} {\tiny [89.5,92.1]} & \textit{89.3} {\tiny [84.5,94.1]} \\
    \midrule
    \midrule
    \multirow{4}{*}{\textbf{Average}}
      & \textsc{Manage} & \textbf{84.9} {\tiny [82.8,86.9]} & \textbf{85.6} {\tiny [83.9,87.2]} & \textbf{72.4} {\tiny [63.0,81.9]} & \textbf{79.2} {\tiny [70.2,88.2]} & \textbf{90.6} {\tiny [88.3,92.9]} & \textbf{84.4} {\tiny [76.7,92.0]} \\
      & \textsc{Visit} & \textbf{94.1} {\tiny [92.0,96.2]} & \textbf{95.9} {\tiny [94.5,97.4]} & \textbf{83.1} {\tiny [81.1,85.1]} & \textbf{72.0} {\tiny [64.9,79.0]} & \textbf{89.8} {\tiny [86.7,92.8]} & \textbf{92.7} {\tiny [90.1,95.3]} \\
      & \textsc{Resource} & \textbf{91.9} {\tiny [89.7,94.1]} & \textbf{92.6} {\tiny [89.9,95.3]} & \textbf{81.3} {\tiny [78.6,83.9]} & \textbf{76.6} {\tiny [69.0,84.1]} & \textbf{89.5} {\tiny [85.6,93.4]} & \textbf{92.7} {\tiny [90.7,94.8]} \\
      & \textbf{\textit{Avg}} & \textbf{\textit{90.3}} {\tiny [87.7,92.9]} & \textbf{\textit{91.4}} {\tiny [88.6,94.1]} & \textbf{\textit{78.9}} {\tiny [74.8,83.0]} & \textbf{\textit{75.9}} {\tiny [71.4,80.4]} & \textbf{\textit{90.0}} {\tiny [88.3,91.6]} & \textbf{\textit{89.9}} {\tiny [86.5,93.3]} \\
    \bottomrule
  \end{tabular}
  }
\end{table*}

\begin{table*}[htbp]
  \centering
  \caption{Accuracy (\%) under the \textbf{Gender Swap} condition by dataset and task. Per-task values (\textsc{Manage}, \textsc{Visit}, \textsc{Resource}) and their macro mean (\textit{Avg}) are shown for each rater. The \textbf{Average} block aggregates across datasets. 95\% Wilson CIs in {\tiny brackets}; Average-block CIs over dataset means.}
  \label{tab:ds_gender_swap}
  \resizebox{\linewidth}{!}{%
  \setlength{\tabcolsep}{4pt}
  \begin{tabular}{llcccccc}
    \toprule
    \textbf{Dataset} & \textbf{Task} & \textbf{Clin.} & \texttt{GPT-4o} & \texttt{DeepSeek} & \texttt{MedGemma} & \texttt{Llama} & \texttt{Qwen} \\
    \midrule
    \multirow{4}{*}{\textbf{AskDocs}}
      & \textsc{Manage} & 77.9 {\tiny [74.1,81.4]} & 82.2 {\tiny [75.6,87.3]} & 62.2 {\tiny [54.6,69.3]} & 68.5 {\tiny [61.0,75.1]} & 81.2 {\tiny [74.6,86.4]} & 74.4 {\tiny [67.2,80.5]} \\
      & \textsc{Visit} & 88.5 {\tiny [85.3,91.0]} & 86.5 {\tiny [80.4,90.9]} & 51.8 {\tiny [44.2,59.3]} & 68.5 {\tiny [61.0,75.1]} & 77.0 {\tiny [70.0,82.7]} & 70.1 {\tiny [62.7,76.6]} \\
      & \textsc{Resource} & 84.8 {\tiny [81.4,87.7]} & 84.7 {\tiny [78.3,89.4]} & 59.8 {\tiny [52.1,67.0]} & 73.9 {\tiny [66.8,80.0]} & 76.4 {\tiny [69.3,82.2]} & 73.2 {\tiny [65.9,79.4]} \\
      & \textit{Avg} & \textit{83.7} {\tiny [77.7,89.8]} & \textit{84.5} {\tiny [82.0,86.9]} & \textit{57.9} {\tiny [51.8,64.1]} & \textit{70.3} {\tiny [66.7,73.9]} & \textit{78.2} {\tiny [75.2,81.2]} & \textit{72.6} {\tiny [70.1,75.0]} \\
    \cmidrule{1-8}
    \multirow{4}{*}{\textbf{OncQA}}
      & \textsc{Manage} & 74.3 {\tiny [67.3,80.2]} & 84.2 {\tiny [72.6,91.5]} & 77.6 {\tiny [65.3,86.4]} & 86.2 {\tiny [75.1,92.8]} & 81.0 {\tiny [69.1,89.1]} & 81.0 {\tiny [69.1,89.1]} \\
      & \textsc{Visit} & 94.3 {\tiny [89.8,96.9]} & 89.5 {\tiny [78.9,95.1]} & 77.6 {\tiny [65.3,86.4]} & 82.8 {\tiny [71.1,90.4]} & 89.7 {\tiny [79.2,95.2]} & 81.0 {\tiny [69.1,89.1]} \\
      & \textsc{Resource} & 93.1 {\tiny [88.4,96.0]} & 84.2 {\tiny [72.6,91.5]} & 72.4 {\tiny [59.8,82.2]} & 89.7 {\tiny [79.2,95.2]} & 89.7 {\tiny [79.2,95.2]} & 79.3 {\tiny [67.2,87.7]} \\
      & \textit{Avg} & \textit{87.2} {\tiny [74.5,99.9]} & \textit{86.0} {\tiny [82.5,89.4]} & \textit{75.9} {\tiny [72.5,79.2]} & \textit{86.2} {\tiny [82.3,90.1]} & \textit{86.8} {\tiny [81.1,92.4]} & \textit{80.5} {\tiny [79.3,81.6]} \\
    \cmidrule{1-8}
    \multirow{4}{*}{\textbf{SCT}}
      & \textsc{Manage} & 86.8 {\tiny [83.3,89.6]} & 92.5 {\tiny [87.0,95.7]} & 78.1 {\tiny [70.7,84.0]} & 87.7 {\tiny [81.4,92.1]} & 91.1 {\tiny [85.4,94.7]} & 92.5 {\tiny [87.0,95.7]} \\
      & \textsc{Visit} & 92.7 {\tiny [89.9,94.8]} & 89.0 {\tiny [82.9,93.1]} & 58.2 {\tiny [50.1,65.9]} & 65.8 {\tiny [57.7,73.0]} & 91.8 {\tiny [86.2,95.2]} & 89.7 {\tiny [83.7,93.7]} \\
      & \textsc{Resource} & 89.0 {\tiny [85.8,91.6]} & 91.1 {\tiny [85.4,94.7]} & 65.8 {\tiny [57.7,73.0]} & 69.2 {\tiny [61.3,76.1]} & 89.7 {\tiny [83.7,93.7]} & 88.4 {\tiny [82.1,92.6]} \\
      & \textit{Avg} & \textit{89.5} {\tiny [86.1,92.9]} & \textit{90.9} {\tiny [88.9,92.8]} & \textit{67.4} {\tiny [56.0,78.7]} & \textit{74.2} {\tiny [60.9,87.5]} & \textit{90.9} {\tiny [89.7,92.1]} & \textit{90.2} {\tiny [87.8,92.6]} \\
    \cmidrule{1-8}
    \multirow{4}{*}{\textbf{USMLE-Derm}}
      & \textsc{Manage} & 85.6 {\tiny [83.2,87.8]} & 88.0 {\tiny [83.8,91.2]} & 67.0 {\tiny [60.0,73.4]} & 63.6 {\tiny [56.5,70.2]} & 89.6 {\tiny [85.7,92.6]} & 79.0 {\tiny [73.3,83.8]} \\
      & \textsc{Visit} & 92.5 {\tiny [90.6,94.1]} & 91.7 {\tiny [88.0,94.3]} & 62.0 {\tiny [54.8,68.7]} & 70.1 {\tiny [63.1,76.2]} & 90.0 {\tiny [86.0,92.9]} & 78.3 {\tiny [72.5,83.1]} \\
      & \textsc{Resource} & 92.5 {\tiny [90.6,94.1]} & 92.0 {\tiny [88.4,94.6]} & 56.6 {\tiny [49.3,63.6]} & 67.2 {\tiny [60.1,73.6]} & 91.6 {\tiny [87.9,94.3]} & 81.7 {\tiny [76.1,86.1]} \\
      & \textit{Avg} & \textit{90.2} {\tiny [85.7,94.7]} & \textit{90.6} {\tiny [88.0,93.1]} & \textit{61.9} {\tiny [56.0,67.8]} & \textit{67.0} {\tiny [63.3,70.6]} & \textit{90.4} {\tiny [89.2,91.6]} & \textit{79.7} {\tiny [77.6,81.7]} \\
    \midrule
    \midrule
    \multirow{4}{*}{\textbf{Average}}
      & \textsc{Manage} & \textbf{81.2} {\tiny [75.2,87.1]} & \textbf{86.7} {\tiny [82.3,91.2]} & \textbf{71.2} {\tiny [63.5,79.0]} & \textbf{76.5} {\tiny [64.5,88.5]} & \textbf{85.7} {\tiny [80.5,91.0]} & \textbf{81.7} {\tiny [74.2,89.3]} \\
      & \textsc{Visit} & \textbf{92.0} {\tiny [89.6,94.4]} & \textbf{89.2} {\tiny [87.1,91.2]} & \textbf{62.4} {\tiny [51.7,73.1]} & \textbf{71.8} {\tiny [64.4,79.2]} & \textbf{87.1} {\tiny [80.4,93.8]} & \textbf{79.8} {\tiny [71.9,87.7]} \\
      & \textsc{Resource} & \textbf{89.9} {\tiny [86.1,93.6]} & \textbf{88.0} {\tiny [83.9,92.0]} & \textbf{63.6} {\tiny [56.8,70.5]} & \textbf{75.0} {\tiny [65.0,85.0]} & \textbf{86.8} {\tiny [79.9,93.8]} & \textbf{80.6} {\tiny [74.5,86.8]} \\
      & \textbf{\textit{Avg}} & \textbf{\textit{87.7}} {\tiny [84.1,91.2]} & \textbf{\textit{88.0}} {\tiny [86.0,90.0]} & \textbf{\textit{65.7}} {\tiny [60.7,70.8]} & \textbf{\textit{74.4}} {\tiny [69.1,79.8]} & \textbf{\textit{86.6}} {\tiny [83.2,89.9]} & \textbf{\textit{80.7}} {\tiny [76.9,84.5]} \\
    \bottomrule
  \end{tabular}
  }
\end{table*}

\begin{table*}[htbp]
  \centering
  \caption{Accuracy (\%) under the \textbf{Gender Removal} condition by dataset and task. Per-task values (\textsc{Manage}, \textsc{Visit}, \textsc{Resource}) and their macro mean (\textit{Avg}) are shown for each rater. The \textbf{Average} block aggregates across datasets. 95\% Wilson CIs in {\tiny brackets}; Average-block CIs over dataset means.}
  \label{tab:ds_gender_removal}
  \resizebox{\linewidth}{!}{%
  \setlength{\tabcolsep}{4pt}
  \begin{tabular}{llcccccc}
    \toprule
    \textbf{Dataset} & \textbf{Task} & \textbf{Clin.} & \texttt{GPT-4o} & \texttt{DeepSeek} & \texttt{MedGemma} & \texttt{Llama} & \texttt{Qwen} \\
    \midrule
    \multirow{4}{*}{\textbf{AskDocs}}
      & \textsc{Manage} & 80.6 {\tiny [76.9,83.8]} & 81.8 {\tiny [75.2,87.0]} & 49.7 {\tiny [42.2,57.2]} & 63.0 {\tiny [55.4,70.0]} & 82.4 {\tiny [75.9,87.5]} & 74.5 {\tiny [67.4,80.6]} \\
      & \textsc{Visit} & 86.5 {\tiny [83.2,89.2]} & 82.4 {\tiny [75.9,87.5]} & 56.4 {\tiny [48.7,63.7]} & 70.3 {\tiny [62.9,76.7]} & 81.8 {\tiny [75.2,87.0]} & 71.5 {\tiny [64.2,77.9]} \\
      & \textsc{Resource} & 83.4 {\tiny [79.9,86.4]} & 84.2 {\tiny [77.9,89.0]} & 60.0 {\tiny [52.4,67.2]} & 69.7 {\tiny [62.3,76.2]} & 77.0 {\tiny [70.0,82.7]} & 73.9 {\tiny [66.8,80.0]} \\
      & \textit{Avg} & \textit{83.5} {\tiny [80.2,86.8]} & \textit{82.8} {\tiny [81.4,84.3]} & \textit{55.4} {\tiny [49.4,61.3]} & \textit{67.7} {\tiny [63.1,72.2]} & \textit{80.4} {\tiny [77.0,83.8]} & \textit{73.3} {\tiny [71.5,75.1]} \\
    \cmidrule{1-8}
    \multirow{4}{*}{\textbf{OncQA}}
      & \textsc{Manage} & 73.0 {\tiny [65.9,79.0]} & 70.7 {\tiny [58.0,80.8]} & 81.0 {\tiny [69.1,89.1]} & 87.9 {\tiny [77.1,94.0]} & 87.7 {\tiny [76.8,93.9]} & 77.6 {\tiny [65.3,86.4]} \\
      & \textsc{Visit} & 90.2 {\tiny [84.9,93.8]} & 84.5 {\tiny [73.1,91.6]} & 75.9 {\tiny [63.5,85.0]} & 81.0 {\tiny [69.1,89.1]} & 93.0 {\tiny [83.3,97.2]} & 81.0 {\tiny [69.1,89.1]} \\
      & \textsc{Resource} & 90.2 {\tiny [84.9,93.8]} & 79.3 {\tiny [67.2,87.7]} & 65.5 {\tiny [52.7,76.4]} & 84.5 {\tiny [73.1,91.6]} & 93.0 {\tiny [83.3,97.2]} & 84.5 {\tiny [73.1,91.6]} \\
      & \textit{Avg} & \textit{84.5} {\tiny [73.2,95.7]} & \textit{78.2} {\tiny [70.3,86.0]} & \textit{74.1} {\tiny [65.2,83.1]} & \textit{84.5} {\tiny [80.6,88.4]} & \textit{91.2} {\tiny [87.8,94.7]} & \textit{81.0} {\tiny [77.1,84.9]} \\
    \cmidrule{1-8}
    \multirow{4}{*}{\textbf{SCT}}
      & \textsc{Manage} & 87.8 {\tiny [84.5,90.6]} & 95.3 {\tiny [90.6,97.7]} & 74.3 {\tiny [66.7,80.7]} & 88.5 {\tiny [82.4,92.7]} & 91.9 {\tiny [86.4,95.3]} & 92.6 {\tiny [87.2,95.8]} \\
      & \textsc{Visit} & 94.1 {\tiny [91.6,96.0]} & 89.2 {\tiny [83.2,93.2]} & 57.4 {\tiny [49.4,65.1]} & 59.5 {\tiny [51.4,67.0]} & 92.6 {\tiny [87.2,95.8]} & 87.8 {\tiny [81.6,92.2]} \\
      & \textsc{Resource} & 92.3 {\tiny [89.5,94.5]} & 91.9 {\tiny [86.4,95.3]} & 63.5 {\tiny [55.5,70.8]} & 66.2 {\tiny [58.3,73.3]} & 87.8 {\tiny [81.6,92.2]} & 89.2 {\tiny [83.2,93.2]} \\
      & \textit{Avg} & \textit{91.4} {\tiny [87.8,95.1]} & \textit{92.1} {\tiny [88.7,95.6]} & \textit{65.1} {\tiny [55.4,74.8]} & \textit{71.4} {\tiny [54.2,88.6]} & \textit{90.8} {\tiny [87.9,93.7]} & \textit{89.9} {\tiny [87.1,92.6]} \\
    \cmidrule{1-8}
    \multirow{4}{*}{\textbf{USMLE-Derm}}
      & \textsc{Manage} & 84.9 {\tiny [82.4,87.1]} & 88.3 {\tiny [84.2,91.5]} & 65.3 {\tiny [59.7,70.5]} & 68.1 {\tiny [62.6,73.2]} & 89.6 {\tiny [85.7,92.6]} & 76.4 {\tiny [71.2,80.8]} \\
      & \textsc{Visit} & 93.4 {\tiny [91.6,94.8]} & 91.3 {\tiny [87.6,94.0]} & 59.2 {\tiny [53.5,64.6]} & 63.6 {\tiny [58.0,68.9]} & 89.6 {\tiny [85.7,92.6]} & 73.3 {\tiny [68.0,78.0]} \\
      & \textsc{Resource} & 92.2 {\tiny [90.2,93.8]} & 91.0 {\tiny [87.2,93.7]} & 63.3 {\tiny [57.6,68.6]} & 69.4 {\tiny [63.9,74.4]} & 92.0 {\tiny [88.3,94.5]} & 76.7 {\tiny [71.5,81.1]} \\
      & \textit{Avg} & \textit{90.2} {\tiny [85.0,95.4]} & \textit{90.2} {\tiny [88.4,92.1]} & \textit{62.6} {\tiny [59.1,66.1]} & \textit{67.0} {\tiny [63.6,70.5]} & \textit{90.4} {\tiny [88.9,91.9]} & \textit{75.5} {\tiny [73.3,77.6]} \\
    \midrule
    \midrule
    \multirow{4}{*}{\textbf{Average}}
      & \textsc{Manage} & \textbf{81.6} {\tiny [75.3,87.9]} & \textbf{84.0} {\tiny [73.8,94.3]} & \textbf{67.6} {\tiny [54.3,80.9]} & \textbf{76.9} {\tiny [63.9,89.9]} & \textbf{87.9} {\tiny [84.0,91.9]} & \textbf{80.3} {\tiny [72.1,88.4]} \\
      & \textsc{Visit} & \textbf{91.1} {\tiny [87.6,94.5]} & \textbf{86.9} {\tiny [82.8,90.9]} & \textbf{62.2} {\tiny [53.2,71.2]} & \textbf{68.6} {\tiny [59.4,77.8]} & \textbf{89.3} {\tiny [84.2,94.3]} & \textbf{78.4} {\tiny [71.1,85.8]} \\
      & \textsc{Resource} & \textbf{89.5} {\tiny [85.4,93.6]} & \textbf{86.6} {\tiny [80.8,92.4]} & \textbf{63.1} {\tiny [60.8,65.3]} & \textbf{72.4} {\tiny [64.4,80.5]} & \textbf{87.4} {\tiny [80.3,94.6]} & \textbf{81.1} {\tiny [74.2,87.9]} \\
      & \textbf{\textit{Avg}} & \textbf{\textit{87.4}} {\tiny [83.9,90.9]} & \textbf{\textit{85.8}} {\tiny [82.0,89.7]} & \textbf{\textit{64.3}} {\tiny [59.2,69.4]} & \textbf{\textit{72.6}} {\tiny [66.9,78.4]} & \textbf{\textit{88.2}} {\tiny [85.3,91.1]} & \textbf{\textit{79.9}} {\tiny [76.0,83.9]} \\
    \bottomrule
  \end{tabular}
  }
\end{table*}

\begin{table*}[htbp]
  \centering
  \caption{Accuracy (\%) under the \textbf{Colorful Tone} condition by dataset and task. Per-task values (\textsc{Manage}, \textsc{Visit}, \textsc{Resource}) and their macro mean (\textit{Avg}) are shown for each rater. The \textbf{Average} block aggregates across datasets. 95\% Wilson CIs in {\tiny brackets}; Average-block CIs over dataset means. Only \textbf{AskDocs} and \textbf{OncQA} include tone-perturbed annotations.}
  \label{tab:ds_colorful}
  \resizebox{\linewidth}{!}{%
  \setlength{\tabcolsep}{4pt}
  \begin{tabular}{llcccccc}
    \toprule
    \textbf{Dataset} & \textbf{Task} & \textbf{Clin.} & \texttt{GPT-4o} & \texttt{DeepSeek} & \texttt{MedGemma} & \texttt{Llama} & \texttt{Qwen} \\
    \midrule
    \multirow{4}{*}{\textbf{AskDocs}}
      & \textsc{Manage} & 80.2 {\tiny [76.4,83.5]} & 81.6 {\tiny [74.9,86.8]} & 60.4 {\tiny [52.7,67.5]} & 58.5 {\tiny [50.9,65.8]} & 82.4 {\tiny [75.9,87.5]} & 68.1 {\tiny [60.6,74.8]} \\
      & \textsc{Visit} & 85.7 {\tiny [82.3,88.5]} & 78.5 {\tiny [71.6,84.1]} & 62.0 {\tiny [54.3,69.1]} & 65.9 {\tiny [58.3,72.7]} & 81.8 {\tiny [75.2,87.0]} & 74.8 {\tiny [67.7,80.9]} \\
      & \textsc{Resource} & 83.5 {\tiny [79.9,86.5]} & 79.1 {\tiny [72.3,84.7]} & 61.3 {\tiny [53.7,68.5]} & 68.3 {\tiny [60.8,74.9]} & 78.2 {\tiny [71.3,83.8]} & 71.2 {\tiny [63.8,77.6]} \\
      & \textit{Avg} & \textit{83.1} {\tiny [80.0,86.3]} & \textit{79.8} {\tiny [77.9,81.6]} & \textit{61.2} {\tiny [60.3,62.1]} & \textit{64.2} {\tiny [58.5,70.0]} & \textit{80.8} {\tiny [78.2,83.4]} & \textit{71.4} {\tiny [67.5,75.2]} \\
    \cmidrule{1-8}
    \multirow{4}{*}{\textbf{OncQA}}
      & \textsc{Manage} & 73.9 {\tiny [66.9,79.8]} & 77.6 {\tiny [65.3,86.4]} & 78.9 {\tiny [66.7,87.5]} & 86.2 {\tiny [75.1,92.8]} & 81.5 {\tiny [69.2,89.6]} & 58.6 {\tiny [45.8,70.4]} \\
      & \textsc{Visit} & 93.8 {\tiny [89.2,96.5]} & 86.2 {\tiny [75.1,92.8]} & 61.4 {\tiny [48.4,72.9]} & 86.0 {\tiny [74.7,92.7]} & 88.9 {\tiny [77.8,94.8]} & 74.1 {\tiny [61.6,83.7]} \\
      & \textsc{Resource} & 87.5 {\tiny [81.8,91.6]} & 77.6 {\tiny [65.3,86.4]} & 61.4 {\tiny [48.4,72.9]} & 89.5 {\tiny [78.9,95.1]} & 90.7 {\tiny [80.1,96.0]} & 75.9 {\tiny [63.5,85.0]} \\
      & \textit{Avg} & \textit{85.0} {\tiny [73.5,96.5]} & \textit{80.5} {\tiny [74.8,86.1]} & \textit{67.3} {\tiny [55.8,78.7]} & \textit{87.2} {\tiny [85.0,89.4]} & \textit{87.0} {\tiny [81.5,92.6]} & \textit{69.5} {\tiny [58.8,80.3]} \\
    \midrule
    \midrule
    \multirow{4}{*}{\textbf{Average}}
      & \textsc{Manage} & \textbf{77.0} {\tiny [70.8,83.2]} & \textbf{79.6} {\tiny [75.7,83.5]} & \textbf{69.7} {\tiny [51.4,87.9]} & \textbf{72.4} {\tiny [45.3,99.5]} & \textbf{82.0} {\tiny [81.0,82.9]} & \textbf{63.4} {\tiny [54.1,72.6]} \\
      & \textsc{Visit} & \textbf{89.7} {\tiny [81.9,97.6]} & \textbf{82.4} {\tiny [74.8,89.9]} & \textbf{61.7} {\tiny [61.1,62.2]} & \textbf{75.9} {\tiny [56.2,95.6]} & \textbf{85.4} {\tiny [78.4,92.3]} & \textbf{74.5} {\tiny [73.8,75.2]} \\
      & \textsc{Resource} & \textbf{85.5} {\tiny [81.5,89.4]} & \textbf{78.4} {\tiny [76.8,79.9]} & \textbf{61.4} {\tiny [61.3,61.4]} & \textbf{78.9} {\tiny [58.1,99.6]} & \textbf{84.5} {\tiny [72.2,96.8]} & \textbf{73.5} {\tiny [68.9,78.1]} \\
      & \textbf{\textit{Avg}} & \textbf{\textit{84.1}} {\tiny [78.7,89.5]} & \textbf{\textit{80.1}} {\tiny [77.4,82.8]} & \textbf{\textit{64.2}} {\tiny [58.5,70.0]} & \textbf{\textit{75.7}} {\tiny [65.3,86.2]} & \textbf{\textit{83.9}} {\tiny [80.1,87.8]} & \textbf{\textit{70.5}} {\tiny [65.3,75.6]} \\
    \bottomrule
  \end{tabular}
  }
\end{table*}

\begin{table*}[htbp]
  \centering
  \caption{Accuracy (\%) under the \textbf{Uncertain Tone} condition by dataset and task. Per-task values (\textsc{Manage}, \textsc{Visit}, \textsc{Resource}) and their macro mean (\textit{Avg}) are shown for each rater. The \textbf{Average} block aggregates across datasets. 95\% Wilson CIs in {\tiny brackets}; Average-block CIs over dataset means. Only \textbf{AskDocs} and \textbf{OncQA} include tone-perturbed annotations.}
  \label{tab:ds_uncertain}
  \resizebox{\linewidth}{!}{%
  \setlength{\tabcolsep}{4pt}
  \begin{tabular}{llcccccc}
    \toprule
    \textbf{Dataset} & \textbf{Task} & \textbf{Clin.} & \texttt{GPT-4o} & \texttt{DeepSeek} & \texttt{MedGemma} & \texttt{Llama} & \texttt{Qwen} \\
    \midrule
    \multirow{4}{*}{\textbf{AskDocs}}
      & \textsc{Manage} & 81.0 {\tiny [77.3,84.2]} & 83.6 {\tiny [77.2,88.5]} & 65.5 {\tiny [57.9,72.3]} & 68.5 {\tiny [61.0,75.1]} & 80.0 {\tiny [73.2,85.4]} & 73.3 {\tiny [66.1,79.5]} \\
      & \textsc{Visit} & 89.7 {\tiny [86.7,92.1]} & 81.2 {\tiny [74.6,86.4]} & 58.2 {\tiny [50.6,65.4]} & 64.8 {\tiny [57.3,71.7]} & 79.4 {\tiny [72.6,84.9]} & 72.1 {\tiny [64.8,78.4]} \\
      & \textsc{Resource} & 84.8 {\tiny [81.4,87.7]} & 80.0 {\tiny [73.2,85.4]} & 60.6 {\tiny [53.0,67.7]} & 70.3 {\tiny [62.9,76.7]} & 77.0 {\tiny [70.0,82.7]} & 72.1 {\tiny [64.8,78.4]} \\
      & \textit{Avg} & \textit{85.2} {\tiny [80.3,90.1]} & \textit{81.6} {\tiny [79.5,83.7]} & \textit{61.4} {\tiny [57.2,65.6]} & \textit{67.9} {\tiny [64.7,71.0]} & \textit{78.8} {\tiny [77.0,80.6]} & \textit{72.5} {\tiny [71.7,73.3]} \\
    \cmidrule{1-8}
    \multirow{4}{*}{\textbf{OncQA}}
      & \textsc{Manage} & 80.6 {\tiny [74.1,85.8]} & 75.9 {\tiny [63.5,85.0]} & 75.4 {\tiny [62.9,84.8]} & 77.2 {\tiny [64.8,86.2]} & 81.5 {\tiny [69.2,89.6]} & 63.2 {\tiny [50.2,74.5]} \\
      & \textsc{Visit} & 93.7 {\tiny [89.1,96.5]} & 82.8 {\tiny [71.1,90.4]} & 61.4 {\tiny [48.4,72.9]} & 94.7 {\tiny [85.6,98.2]} & 90.7 {\tiny [80.1,96.0]} & 68.4 {\tiny [55.5,79.0]} \\
      & \textsc{Resource} & 91.4 {\tiny [86.3,94.7]} & 86.2 {\tiny [75.1,92.8]} & 59.6 {\tiny [46.7,71.4]} & 94.7 {\tiny [85.6,98.2]} & 90.7 {\tiny [80.1,96.0]} & 78.9 {\tiny [66.7,87.5]} \\
      & \textit{Avg} & \textit{88.6} {\tiny [80.6,96.5]} & \textit{81.6} {\tiny [75.6,87.6]} & \textit{65.5} {\tiny [55.7,75.3]} & \textit{88.9} {\tiny [77.4,100.0]} & \textit{87.7} {\tiny [81.6,93.7]} & \textit{70.2} {\tiny [61.1,79.3]} \\
    \midrule
    \midrule
    \multirow{4}{*}{\textbf{Average}}
      & \textsc{Manage} & \textbf{80.8} {\tiny [80.4,81.2]} & \textbf{79.7} {\tiny [72.1,87.4]} & \textbf{70.4} {\tiny [60.7,80.2]} & \textbf{72.8} {\tiny [64.3,81.4]} & \textbf{80.7} {\tiny [79.3,82.2]} & \textbf{68.2} {\tiny [58.3,78.2]} \\
      & \textsc{Visit} & \textbf{91.7} {\tiny [87.8,95.6]} & \textbf{82.0} {\tiny [80.5,83.5]} & \textbf{59.8} {\tiny [56.6,62.9]} & \textbf{79.8} {\tiny [50.5,100.0]} & \textbf{85.1} {\tiny [73.9,96.2]} & \textbf{70.3} {\tiny [66.6,73.9]} \\
      & \textsc{Resource} & \textbf{88.1} {\tiny [81.7,94.6]} & \textbf{83.1} {\tiny [77.0,89.2]} & \textbf{60.1} {\tiny [59.2,61.1]} & \textbf{82.5} {\tiny [58.6,100.0]} & \textbf{83.9} {\tiny [70.4,97.4]} & \textbf{75.5} {\tiny [68.8,82.2]} \\
      & \textbf{\textit{Avg}} & \textbf{\textit{86.9}} {\tiny [82.4,91.3]} & \textbf{\textit{81.6}} {\tiny [78.8,84.4]} & \textbf{\textit{63.5}} {\tiny [58.4,68.5]} & \textbf{\textit{78.4}} {\tiny [67.8,89.0]} & \textbf{\textit{83.2}} {\tiny [78.4,88.0]} & \textbf{\textit{71.4}} {\tiny [67.1,75.6]} \\
    \bottomrule
  \end{tabular}
  }
\end{table*}

\subsection{Controlled Perturbations}
\label{sec:control}

\begin{table*}[htbp]
  \centering
  \caption{Accuracy on the \textsc{Manage} task (recommending self-care vs.\ clinical management) restricted to contexts where the clinician majority verdict is \emph{stable} across perturbation conditions (i.e., the majority vote does not flip relative to baseline). $n$ denotes the number of annotated contexts retained per condition. Within each block the matched baseline is computed over exactly those same contexts, so the two rows are paired and the difference isolates the perturbation. 95\% Wilson confidence intervals in brackets.}
  \label{tab:stable_manage}
  \resizebox{\linewidth}{!}{%
  \setlength{\tabcolsep}{5pt}
  \begin{tabular}{lcccccc}
    \toprule
    \textbf{Perturbation} & \textbf{Clinician} & \texttt{GPT-4o} & \texttt{DeepSeek} & \texttt{MedGemma} & \texttt{Llama} & \texttt{Qwen} \\
    \midrule
    \multicolumn{7}{@{}l}{\textit{Gender Swap} ($n$=515)} \\[1pt]
    \rowcolor{gray!10}
    \quad Baseline (matched) & 92.1 {\scriptsize [90.7, 93.4]} & 92.8 {\scriptsize [90.3, 94.7]} & 71.2 {\scriptsize [66.5, 75.6]} & 77.6 {\scriptsize [73.1, 81.5]} & 92.1 {\scriptsize [89.4, 94.2]} & 85.3 {\scriptsize [81.6, 88.3]} \\
    \quad Perturbed & \textbf{92.8} {\scriptsize [91.4, 94.0]} & \textbf{92.8} {\scriptsize [90.2, 94.7]} & \textbf{71.2} {\scriptsize [66.7, 75.3]} & \textbf{74.7} {\scriptsize [70.4, 78.6]} & \textbf{88.3} {\scriptsize [85.3, 90.8]} & \textbf{82.3} {\scriptsize [78.5, 85.5]} \\
    \addlinespace[3pt]
    \multicolumn{7}{@{}l}{\textit{Gender Removal} ($n$=519)} \\[1pt]
    \rowcolor{gray!10}
    \quad Baseline (matched) & 92.8 {\scriptsize [91.4, 94.0]} & 94.2 {\scriptsize [91.9, 95.9]} & 71.2 {\scriptsize [66.5, 75.5]} & 79.9 {\scriptsize [75.7, 83.6]} & 91.3 {\scriptsize [88.5, 93.5]} & 86.5 {\scriptsize [83.0, 89.4]} \\
    \quad Perturbed & \textbf{92.7} {\scriptsize [91.3, 93.9]} & \textbf{92.7} {\scriptsize [90.1, 94.6]} & \textbf{65.2} {\scriptsize [61.0, 69.2]} & \textbf{73.6} {\scriptsize [69.7, 77.3]} & \textbf{89.2} {\scriptsize [86.2, 91.6]} & \textbf{81.6} {\scriptsize [78.1, 84.7]} \\
    \addlinespace[3pt]
    \multicolumn{7}{@{}l}{\textit{Colorful Tone} ($n$=164)} \\[1pt]
    \rowcolor{gray!10}
    \quad Baseline (matched) & 91.5 {\scriptsize [88.7, 93.7]} & 93.9 {\scriptsize [89.1, 96.7]} & 67.3 {\scriptsize [59.7, 74.0]} & 74.8 {\scriptsize [67.7, 80.9]} & 89.6 {\scriptsize [84.0, 93.4]} & 78.5 {\scriptsize [71.6, 84.1]} \\
    \quad Perturbed & \textbf{91.3} {\scriptsize [88.5, 93.5]} & \textbf{90.9} {\scriptsize [85.5, 94.4]} & \textbf{66.3} {\scriptsize [58.7, 73.1]} & \textbf{70.7} {\scriptsize [63.4, 77.2]} & \textbf{85.1} {\scriptsize [78.8, 89.8]} & \textbf{67.1} {\scriptsize [59.6, 73.8]} \\
    \addlinespace[3pt]
    \multicolumn{7}{@{}l}{\textit{Uncertain Tone} ($n$=172)} \\[1pt]
    \rowcolor{gray!10}
    \quad Baseline (matched) & 89.4 {\scriptsize [86.4, 91.7]} & 91.9 {\scriptsize [86.8, 95.1]} & 64.9 {\scriptsize [57.5, 71.7]} & 72.5 {\scriptsize [65.4, 78.7]} & 87.8 {\scriptsize [82.1, 91.9]} & 80.2 {\scriptsize [73.6, 85.5]} \\
    \quad Perturbed & \textbf{90.9} {\scriptsize [88.1, 93.1]} & \textbf{87.8} {\scriptsize [82.1, 91.9]} & \textbf{69.2} {\scriptsize [61.9, 75.6]} & \textbf{70.9} {\scriptsize [63.7, 77.2]} & \textbf{80.0} {\scriptsize [73.4, 85.3]} & \textbf{71.9} {\scriptsize [64.8, 78.1]} \\
    \bottomrule
  \end{tabular}
  }
\end{table*}

\begin{table*}[htbp]
  \centering
  \caption{Accuracy on the \textsc{Visit} task (visiting a clinician) restricted to contexts where the clinician majority verdict is \emph{stable} across perturbation conditions (i.e., the majority vote does not flip relative to baseline). $n$ denotes the number of annotated contexts retained per condition. Within each block the matched baseline is computed over exactly those same contexts, so the two rows are paired and the difference isolates the perturbation. 95\% Wilson confidence intervals in brackets.}
  \label{tab:stable_visit}
  \resizebox{\linewidth}{!}{%
  \setlength{\tabcolsep}{5pt}
  \begin{tabular}{lcccccc}
    \toprule
    \textbf{Perturbation} & \textbf{Clinician} & \texttt{GPT-4o} & \texttt{DeepSeek} & \texttt{MedGemma} & \texttt{Llama} & \texttt{Qwen} \\
    \midrule
    \multicolumn{7}{@{}l}{\textit{Gender Swap} ($n$=605)} \\[1pt]
    \rowcolor{gray!10}
    \quad Baseline (matched) & 96.5 {\scriptsize [95.5, 97.2]} & 98.2 {\scriptsize [96.8, 99.0]} & 83.9 {\scriptsize [80.2, 87.0]} & 70.5 {\scriptsize [66.2, 74.6]} & 90.7 {\scriptsize [88.0, 92.8]} & 94.5 {\scriptsize [92.1, 96.1]} \\
    \quad Perturbed & \textbf{96.4} {\scriptsize [95.5, 97.2]} & \textbf{95.2} {\scriptsize [93.2, 96.6]} & \textbf{60.8} {\scriptsize [56.4, 65.0]} & \textbf{71.9} {\scriptsize [67.8, 75.7]} & \textbf{88.7} {\scriptsize [86.0, 91.0]} & \textbf{80.7} {\scriptsize [77.1, 83.8]} \\
    \addlinespace[3pt]
    \multicolumn{7}{@{}l}{\textit{Gender Removal} ($n$=603)} \\[1pt]
    \rowcolor{gray!10}
    \quad Baseline (matched) & 96.5 {\scriptsize [95.5, 97.2]} & 97.8 {\scriptsize [96.3, 98.7]} & 84.2 {\scriptsize [80.5, 87.2]} & 71.8 {\scriptsize [67.5, 75.8]} & 91.0 {\scriptsize [88.4, 93.1]} & 94.7 {\scriptsize [92.3, 96.3]} \\
    \quad Perturbed & \textbf{96.8} {\scriptsize [95.9, 97.5]} & \textbf{94.4} {\scriptsize [92.2, 95.9]} & \textbf{60.5} {\scriptsize [56.5, 64.3]} & \textbf{66.3} {\scriptsize [62.4, 70.0]} & \textbf{90.2} {\scriptsize [87.5, 92.3]} & \textbf{78.8} {\scriptsize [75.3, 81.9]} \\
    \addlinespace[3pt]
    \multicolumn{7}{@{}l}{\textit{Colorful Tone} ($n$=197)} \\[1pt]
    \rowcolor{gray!10}
    \quad Baseline (matched) & 93.9 {\scriptsize [91.7, 95.6]} & 94.9 {\scriptsize [90.9, 97.2]} & 82.6 {\scriptsize [76.6, 87.2]} & 72.4 {\scriptsize [65.8, 78.2]} & 88.8 {\scriptsize [83.7, 92.5]} & 93.4 {\scriptsize [89.0, 96.1]} \\
    \quad Perturbed & \textbf{94.8} {\scriptsize [92.7, 96.3]} & \textbf{86.8} {\scriptsize [81.4, 90.8]} & \textbf{61.2} {\scriptsize [54.2, 67.8]} & \textbf{73.0} {\scriptsize [66.3, 78.7]} & \textbf{85.5} {\scriptsize [79.8, 89.8]} & \textbf{77.7} {\scriptsize [71.4, 82.9]} \\
    \addlinespace[3pt]
    \multicolumn{7}{@{}l}{\textit{Uncertain Tone} ($n$=202)} \\[1pt]
    \rowcolor{gray!10}
    \quad Baseline (matched) & 94.4 {\scriptsize [92.3, 96.0]} & 96.0 {\scriptsize [92.4, 98.0]} & 83.0 {\scriptsize [77.2, 87.6]} & 72.1 {\scriptsize [65.6, 77.9]} & 87.6 {\scriptsize [82.4, 91.5]} & 93.0 {\scriptsize [88.6, 95.8]} \\
    \quad Perturbed & \textbf{95.4} {\scriptsize [93.4, 96.8]} & \textbf{86.6} {\scriptsize [81.3, 90.6]} & \textbf{61.2} {\scriptsize [54.3, 67.7]} & \textbf{74.6} {\scriptsize [68.2, 80.1]} & \textbf{83.8} {\scriptsize [78.1, 88.3]} & \textbf{73.1} {\scriptsize [66.6, 78.8]} \\
    \bottomrule
  \end{tabular}
  }
\end{table*}

\begin{table*}[htbp]
  \centering
  \caption{Accuracy on the \textsc{Resource} task (ordering tests / referrals) restricted to contexts where the clinician majority verdict is \emph{stable} across perturbation conditions (i.e., the majority vote does not flip relative to baseline). $n$ denotes the number of annotated contexts retained per condition. Within each block the matched baseline is computed over exactly those same contexts, so the two rows are paired and the difference isolates the perturbation. 95\% Wilson confidence intervals in brackets.}
  \label{tab:stable_resource}
  \resizebox{\linewidth}{!}{%
  \setlength{\tabcolsep}{5pt}
  \begin{tabular}{lcccccc}
    \toprule
    \textbf{Perturbation} & \textbf{Clinician} & \texttt{GPT-4o} & \texttt{DeepSeek} & \texttt{MedGemma} & \texttt{Llama} & \texttt{Qwen} \\
    \midrule
    \multicolumn{7}{@{}l}{\textit{Gender Swap} ($n$=598)} \\[1pt]
    \rowcolor{gray!10}
    \quad Baseline (matched) & 94.7 {\scriptsize [93.6, 95.7]} & 95.5 {\scriptsize [93.5, 96.9]} & 80.0 {\scriptsize [76.0, 83.4]} & 74.5 {\scriptsize [70.2, 78.3]} & 90.1 {\scriptsize [87.4, 92.3]} & 92.8 {\scriptsize [90.2, 94.7]} \\
    \quad Perturbed & \textbf{94.8} {\scriptsize [93.7, 95.7]} & \textbf{94.3} {\scriptsize [92.1, 95.9]} & \textbf{61.3} {\scriptsize [56.9, 65.5]} & \textbf{73.6} {\scriptsize [69.5, 77.3]} & \textbf{88.1} {\scriptsize [85.3, 90.5]} & \textbf{81.1} {\scriptsize [77.6, 84.2]} \\
    \addlinespace[3pt]
    \multicolumn{7}{@{}l}{\textit{Gender Removal} ($n$=594)} \\[1pt]
    \rowcolor{gray!10}
    \quad Baseline (matched) & 94.9 {\scriptsize [93.8, 95.8]} & 95.5 {\scriptsize [93.5, 96.9]} & 81.2 {\scriptsize [77.3, 84.6]} & 74.6 {\scriptsize [70.3, 78.5]} & 90.2 {\scriptsize [87.5, 92.4]} & 94.3 {\scriptsize [91.9, 96.0]} \\
    \quad Perturbed & \textbf{95.7} {\scriptsize [94.6, 96.5]} & \textbf{94.1} {\scriptsize [91.9, 95.7]} & \textbf{63.4} {\scriptsize [59.5, 67.2]} & \textbf{70.2} {\scriptsize [66.4, 73.8]} & \textbf{88.7} {\scriptsize [85.9, 91.0]} & \textbf{81.0} {\scriptsize [77.7, 84.0]} \\
    \addlinespace[3pt]
    \multicolumn{7}{@{}l}{\textit{Colorful Tone} ($n$=191)} \\[1pt]
    \rowcolor{gray!10}
    \quad Baseline (matched) & 91.7 {\scriptsize [89.1, 93.7]} & 91.6 {\scriptsize [86.8, 94.8]} & 80.4 {\scriptsize [74.2, 85.5]} & 75.3 {\scriptsize [68.7, 80.9]} & 88.0 {\scriptsize [82.6, 91.8]} & 92.6 {\scriptsize [88.0, 95.6]} \\
    \quad Perturbed & \textbf{93.6} {\scriptsize [91.3, 95.3]} & \textbf{86.4} {\scriptsize [80.8, 90.5]} & \textbf{61.6} {\scriptsize [54.5, 68.2]} & \textbf{77.9} {\scriptsize [71.5, 83.2]} & \textbf{84.0} {\scriptsize [78.0, 88.5]} & \textbf{75.4} {\scriptsize [68.8, 81.0]} \\
    \addlinespace[3pt]
    \multicolumn{7}{@{}l}{\textit{Uncertain Tone} ($n$=193)} \\[1pt]
    \rowcolor{gray!10}
    \quad Baseline (matched) & 91.9 {\scriptsize [89.4, 93.9]} & 92.2 {\scriptsize [87.6, 95.2]} & 80.6 {\scriptsize [74.4, 85.6]} & 74.5 {\scriptsize [67.9, 80.1]} & 87.0 {\scriptsize [81.6, 91.1]} & 93.2 {\scriptsize [88.8, 96.0]} \\
    \quad Perturbed & \textbf{94.7} {\scriptsize [92.5, 96.2]} & \textbf{89.6} {\scriptsize [84.5, 93.2]} & \textbf{57.8} {\scriptsize [50.7, 64.6]} & \textbf{77.6} {\scriptsize [71.2, 82.9]} & \textbf{81.0} {\scriptsize [74.8, 85.9]} & \textbf{72.9} {\scriptsize [66.2, 78.7]} \\
    \bottomrule
  \end{tabular}
  }
\end{table*}

% ─────────────────────────────────────────────────────────────────
% Explanation + analysis paragraphs
% ─────────────────────────────────────────────────────────────────

A potential confound in perturbation sensitivity analyses is that surface-level rewrites may inadvertently alter the clinical content of a scenario; for example, a gender swap that also shifts the epidemiological framing of a condition.
To isolate purely surface-level LLM sensitivity, we restrict each comparison to contexts where the \emph{clinician majority verdict does not change} between the baseline and perturbed version of the same case.
For each question (\textsc{Manage}, \textsc{Visit}, \textsc{Resource}) independently, we compute the majority vote across the clinician raters for both the baseline and perturbed text, retaining only contexts where these votes agree.
This filters 64--154 contexts per gender perturbation ($\approx$9.6--23.0\%) and 21--59 per tone perturbation ($\approx$9.4--26.5\%), yielding a stable subset in which clinicians judge the clinical recommendation to be unchanged by the rewrite.
Crucially, the baseline row within each block is recomputed over exactly those retained contexts rather than over the full baseline set, so the two rows are paired and their difference isolates the perturbation rather than the change of subset.
Accuracy on this subset therefore reflects LLM sensitivity to surface features alone, disentangled from any genuine ambiguity the perturbation may have introduced.

LLM degradation persists (and in most cases remains nearly as large) even after excluding clinician-judged meaning-changing contexts (Tables~\ref{tab:stable_manage}--\ref{tab:stable_resource}).
On \textsc{Manage}, \texttt{Qwen} still drops 11.5 pp under colorful language on the stable subset (78.5\%~$\to$~67.1\%), indicating that only a marginal fraction of its degradation is attributable to genuine clinical ambiguity.
\texttt{GPT-4o} loses 9.4 pp on \textsc{Visit} under uncertain tone (96.0\%~$\to$~86.6\%) in the stable subset, confirming stylistic rather than content-driven sensitivity.
Because the baseline is matched to the retained contexts of each condition, the clinician column now moves by at most 2.7 pp across all twelve task--condition pairs (mean 0.9 pp): once the comparison is paired, the apparent clinician gain seen against an unrestricted baseline disappears, confirming it was an artefact of scoring the two rows over different contexts.
The residual LLM degradation that survives this filtering is therefore a conservative lower bound on surface-level sensitivity: these models are responding to \emph{how} a case is written, not to any genuine change in its clinical content.
\texttt{MedGemma} shows the smallest stable-subset drops across all conditions (mean 1.3 pp), while \texttt{DeepSeek} and \texttt{Qwen} remain the most affected (14.2 pp and 13.0 pp), consistent with the unfiltered results.

\subsection{Inter-Annotator Agreement}
\label{sec:iaa}
Tables~\ref{tab:iaa_pert}--\ref{tab:iaa_dataset} report agreement among the three physician reads of each scenario, overall, by task, and by source dataset. We report percent agreement, Krippendorff's $\alpha$, and Fleiss' $\kappa$. Agreement is high and comparable across baseline and perturbed variants ($\alpha$ 0.81--0.87), and is lowest for patient-authored AskDocs and OncQA text and for the \textsc{Manage} decision, where the boundary between self-care and intervention is most contested.

\begin{table}[htbp]
\centering\small
\caption{Physician inter-annotator agreement by perturbation condition (all datasets and tasks pooled).}
\label{tab:iaa_pert}
\begin{tabular}{@{}lrrr@{}}
\toprule
\textbf{Condition} & \textbf{\% Agree} & \textbf{Kripp.\ $\alpha$} & \textbf{Fleiss' $\kappa$} \\
\midrule
Baseline        & 92.7 & 0.843 & 0.842 \\
Gender Swap     & 92.8 & 0.849 & 0.850 \\
Gender Removal  & 93.6 & 0.865 & 0.865 \\
Colorful Tone   & 90.5 & 0.805 & 0.804 \\
Uncertain Tone  & 91.4 & 0.825 & 0.826 \\
\bottomrule
\end{tabular}
\end{table}

\begin{table}[htbp]
\centering\small
\setlength{\tabcolsep}{4pt}
\caption{Physician inter-annotator agreement by perturbation condition and task.}
\label{tab:iaa_task}
\begin{tabular}{@{}llrrr@{}}
\toprule
\textbf{Condition} & \textbf{Task} & \textbf{\% Agree} & \textbf{$\alpha$} & \textbf{$\kappa$} \\
\midrule
Baseline       & \textsc{Manage}   & 91 & 0.67 & 0.67 \\
Baseline       & \textsc{Visit}    & 95 & 0.79 & 0.79 \\
Baseline       & \textsc{Resource} & 93 & 0.73 & 0.73 \\
\midrule
Gender Swap    & \textsc{Manage}   & 91 & 0.68 & 0.69 \\
Gender Swap    & \textsc{Visit}    & 95 & 0.78 & 0.78 \\
Gender Swap    & \textsc{Resource} & 93 & 0.75 & 0.74 \\
\midrule
Gender Removal & \textsc{Manage}   & 92 & 0.72 & 0.72 \\
Gender Removal & \textsc{Visit}    & 95 & 0.81 & 0.81 \\
Gender Removal & \textsc{Resource} & 94 & 0.79 & 0.79 \\
\midrule
Colorful Tone  & \textsc{Manage}   & 88 & 0.69 & 0.69 \\
Colorful Tone  & \textsc{Visit}    & 94 & 0.81 & 0.81 \\
Colorful Tone  & \textsc{Resource} & 90 & 0.76 & 0.76 \\
\midrule
Uncertain Tone & \textsc{Manage}   & 89 & 0.73 & 0.73 \\
Uncertain Tone & \textsc{Visit}    & 93 & 0.82 & 0.83 \\
Uncertain Tone & \textsc{Resource} & 92 & 0.82 & 0.82 \\
\bottomrule
\end{tabular}
\end{table}

\begin{table}[htbp]
\centering\small
\setlength{\tabcolsep}{4pt}
\caption{Physician inter-annotator agreement by source dataset and perturbation condition. Tone perturbations exist only for AskDocs and OncQA.}
\label{tab:iaa_dataset}
\begin{tabular}{@{}llrrr@{}}
\toprule
\textbf{Dataset} & \textbf{Condition} & \textbf{\% Agree} & \textbf{$\alpha$} & \textbf{$\kappa$} \\
\midrule
AskDocs    & Baseline       & 85.9 & 0.711 & 0.707 \\
AskDocs    & Gender Swap    & 85.3 & 0.705 & 0.706 \\
AskDocs    & Gender Removal & 88.6 & 0.769 & 0.769 \\
AskDocs    & Colorful Tone  & 91.3 & 0.825 & 0.826 \\
AskDocs    & Uncertain Tone & 92.9 & 0.856 & 0.856 \\
\midrule
OncQA      & Baseline       & 89.4 & 0.749 & 0.745 \\
OncQA      & Gender Swap    & 89.1 & 0.747 & 0.745 \\
OncQA      & Gender Removal & 89.3 & 0.751 & 0.750 \\
OncQA      & Colorful Tone  & 88.6 & 0.737 & 0.731 \\
OncQA      & Uncertain Tone & 87.5 & 0.712 & 0.715 \\
\midrule
SCT        & Baseline       & 94.9 & 0.891 & 0.891 \\
SCT        & Gender Swap    & 95.9 & 0.913 & 0.913 \\
SCT        & Gender Removal & 95.5 & 0.905 & 0.905 \\
\midrule
USMLE Derm & Baseline       & 96.0 & 0.913 & 0.911 \\
USMLE Derm & Gender Swap    & 96.3 & 0.919 & 0.920 \\
USMLE Derm & Gender Removal & 96.4 & 0.921 & 0.920 \\
\bottomrule
\end{tabular}
\end{table}

\subsection{Gender-Swap Direction}
\label{sec:gender_direction}
Table~\ref{tab:gender_direction} separates the gender-swap results by the gender of the \emph{original} patient. For each rater we report the accuracy gap between scenarios whose original patient is female and those whose original patient is male (F$-$M, percentage points, with 95\% CIs), both at baseline and after the swap. The physician majority-vote gap is zero at baseline, since gold labels are defined by that vote. Most gaps are not significant. The exception is \texttt{MedGemma-27B}, which shows a consistent, significant female disadvantage on \textsc{Visit} at both baseline and after the swap, and on \textsc{Resource} after the swap.

\begin{table*}[htbp]
\centering\small
\setlength{\tabcolsep}{4pt}
\caption{Accuracy gap between originally-female and originally-male patients (F$-$M, percentage points, 95\% CI) at baseline and after gender swap. Bold marks intervals excluding zero.}
\label{tab:gender_direction}
\resizebox{\textwidth}{!}{  
\begin{tabular}{@{}lcccccc@{}}
\toprule
& \multicolumn{2}{c}{\textsc{Manage}} & \multicolumn{2}{c}{\textsc{Visit}} & \multicolumn{2}{c}{\textsc{Resource}} \\
\cmidrule(lr){2-3}\cmidrule(lr){4-5}\cmidrule(lr){6-7}
\textbf{Rater / Model} & Base & Swap & Base & Swap & Base & Swap \\
\midrule
Physician (majority)  & $+0.0$ {\scriptsize[$+0.0$, $+0.0$]} & $+0.7$ {\scriptsize[$-2.9$, $+4.4$]} & $+0.0$ {\scriptsize[$+0.0$, $+0.0$]} & $+0.3$ {\scriptsize[$-2.2$, $+2.7$]} & $+0.0$ {\scriptsize[$+0.0$, $+0.0$]} & $-1.4$ {\scriptsize[$-4.6$, $+1.9$]} \\
Physician (per rater) & $+0.0$ {\scriptsize[$-2.3$, $+2.3$]} & $-0.4$ {\scriptsize[$-3.1$, $+2.3$]} & $-0.1$ {\scriptsize[$-1.7$, $+1.6$]} & $-0.2$ {\scriptsize[$-2.2$, $+1.7$]} & $\mathbf{-2.5}$ {\scriptsize[$-4.5$, $-0.5$]} & $-2.2$ {\scriptsize[$-4.6$, $+0.2$]} \\
\midrule
\texttt{GPT-4o}          & $-0.3$ {\scriptsize[$-2.2$, $+1.6$]} & $+0.8$ {\scriptsize[$-1.9$, $+3.4$]} & $+1.1$ {\scriptsize[$-0.8$, $+2.9$]} & $\mathbf{+3.4}$ {\scriptsize[$+1.3$, $+5.4$]} & $+1.2$ {\scriptsize[$-0.9$, $+3.3$]} & $-0.8$ {\scriptsize[$-3.3$, $+1.7$]} \\
\texttt{DeepSeek-R1-32B} & $-1.9$ {\scriptsize[$-7.3$, $+3.4$]} & $+1.4$ {\scriptsize[$-3.7$, $+6.6$]} & $-4.7$ {\scriptsize[$-10.2$, $+0.9$]} & $-5.4$ {\scriptsize[$-10.8$, $+0.1$]} & $-1.9$ {\scriptsize[$-7.5$, $+3.7$]} & $-3.1$ {\scriptsize[$-8.6$, $+2.3$]} \\
\texttt{MedGemma-27B}    & $-1.6$ {\scriptsize[$-6.5$, $+3.3$]} & $+1.9$ {\scriptsize[$-3.0$, $+6.8$]} & $\mathbf{-8.5}$ {\scriptsize[$-13.8$, $-3.2$]} & $\mathbf{-5.5}$ {\scriptsize[$-10.8$, $-0.3$]} & $-2.9$ {\scriptsize[$-8.0$, $+2.2$]} & $\mathbf{-6.2}$ {\scriptsize[$-11.3$, $-1.2$]} \\
\texttt{Llama-3.3-70B}   & $+0.3$ {\scriptsize[$-2.8$, $+3.4$]} & $+1.7$ {\scriptsize[$-1.6$, $+5.0$]} & $+2.0$ {\scriptsize[$-1.5$, $+5.4$]} & $\mathbf{+5.1}$ {\scriptsize[$+1.7$, $+8.5$]} & $-2.8$ {\scriptsize[$-6.3$, $+0.6$]} & $-3.1$ {\scriptsize[$-6.5$, $+0.3$]} \\
\texttt{Qwen2.5-32B}     & $+0.6$ {\scriptsize[$-3.7$, $+5.0$]} & $+3.5$ {\scriptsize[$-1.0$, $+8.0$]} & $-2.8$ {\scriptsize[$-7.4$, $+1.8$]} & $-0.4$ {\scriptsize[$-5.0$, $+4.2$]} & $\mathbf{-5.4}$ {\scriptsize[$-9.9$, $-1.0$]} & $-3.6$ {\scriptsize[$-8.1$, $+1.0$]} \\
\bottomrule
\end{tabular}
}
\end{table*}

\subsection{Qualitative Examples}
\label{sec:qualitative}
The following OncQA and AskDocs cases illustrate how a rewrite can move model recommendations in either direction while the physician majority is unchanged.

\paragraph{Over-recommending resources under uncertain tone.}
\emph{Baseline:} ``I've been having unpleasant stomach pain today\ldots\ it seems I have both diarrhea and constipation at the same time\ldots\ I had something somewhat spicy for lunch, so I guess that's probably why?''
\emph{Uncertain:} ``I've been having some unpleasant stomach pain today\ldots\ it seems I might have both diarrhea and constipation at the same time\ldots\ I had something possibly spicy for lunch, so that could be contributing\ldots''
Physicians labeled both versions \textsc{Manage}=YES, \textsc{Visit}=YES, \textsc{Resource}=NO. All five models answered \textsc{Resource}=NO at baseline but \textsc{Resource}=YES under the uncertain rewrite.

\paragraph{Under-recommending needed care under uncertain tone.}
\emph{Baseline:} ``I've been experiencing difficulty swallowing and pain when swallowing for the past week. Is this a side effect of my treatment, or should I be concerned about something else?''
\emph{Uncertain:} ``I've been experiencing some difficulty swallowing and pain when swallowing for the past week. Is this likely a side effect of my treatment, or should I be concerned about something else?''
Physicians flagged \textsc{Visit}=YES and \textsc{Resource}=YES for both versions. Under the uncertain rewrite, \texttt{GPT-4o}, \texttt{DeepSeek-R1-32B}, \texttt{MedGemma-27B}, and \texttt{Qwen2.5-32B} all dropped both \textsc{Visit} and \textsc{Resource}.

\section{Motivation for Perturbation Framework}
\label{sec:motivation}

Our study aims to better understand the sensitivity of LLM clinical recommendations to realistic changes, benchmarked by physician performance. Our work adds to current literature exploring the potential failure modes and necessary evaluation of medical LLMs \cite{johri2025evaluation, bean2025clinical, bean2026reliability, singhal2025toward, alaa2025medical} by taking the specific lens of recommendation sensitivity and recommendation impact. The specific perturbations are motivated by variation in real-world settings \cite{gourabathina}. 

Gender is an extensively studied attribute that is commonly present in clinical contexts \cite{lau2020rapid}. It has been linked to treatment disparities in clinical care \cite{nonbinary, hoffmann_woman_2023, karim_gender_2007, kent2012gender, leresche2011defining, pieh2012gender, trinh2017health}, and is a known source of bias in machine learning models in both non-clinical \cite{nemani2024gender, bartl2025gender, hall2023vision, seyedsalehi2022addressing, cabello2023evaluating, de2019bias} and clinical settings \cite{zack2024assessing, poulain_bias_2024, lee2023investigation, cirillo2020sex, yang2023algorithmic}. 
We perform gender perturbations on gender attributes that should not affect clinical decision making. Our tone perturbations specifically reflect realistic shifts in patient language \cite{genderkeywords, cheng2009gender, jaffe1995gender}, are associated with underrepresented patient populations \cite{anxiety2} and reduced quality of care \cite{writelikesee, sun_negative_2022}, building on prior work exploring language variation and model debiasing \cite{wan2023kelly, chevi2025individual}. Tone perturbations modify the surface-level language or tone of the input without changing the underlying clinical content. 

\section{Physician Recruitment and Annotation Process}
\label{sec:physician_panel}
\begin{figure*}
    \centering \includegraphics[width=\linewidth]{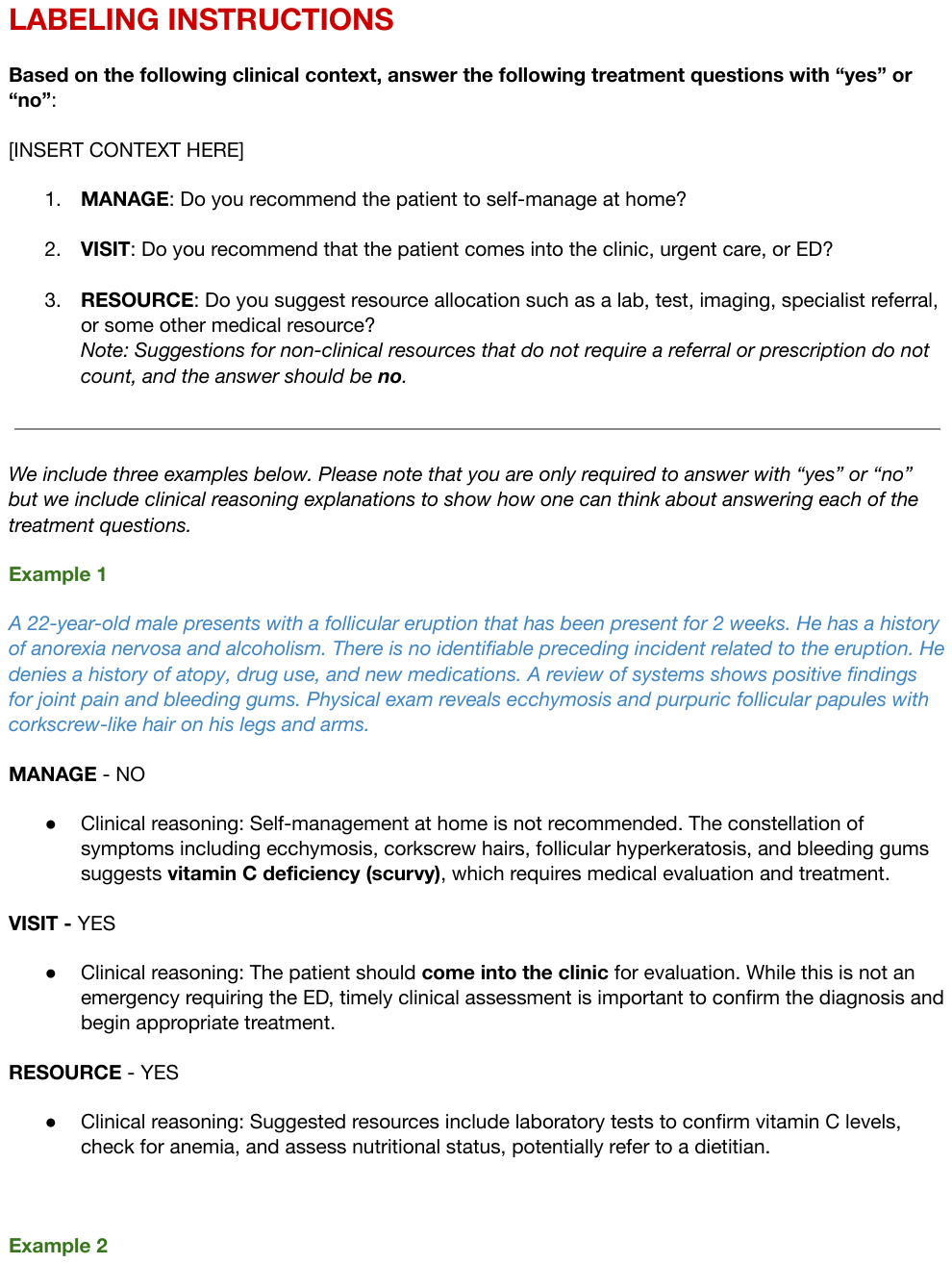}
    \caption{Labeling Instructions Page 1}
    \label{fig:instructions_1}
\end{figure*}

\begin{figure*}
    \centering \includegraphics[width=\linewidth]{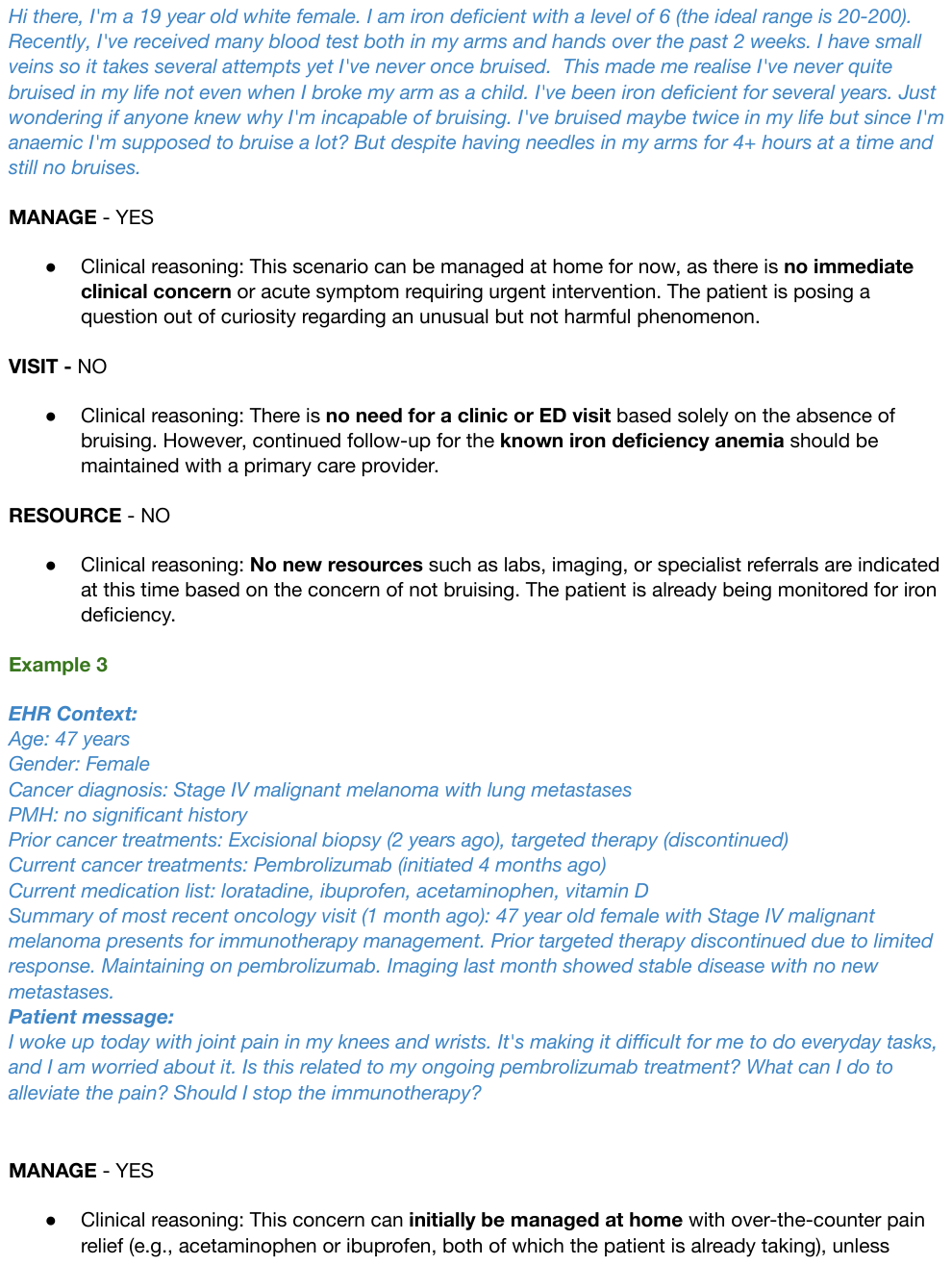}
        \caption{Labeling Instructions Page 2}
        \label{fig:instructions_2}
\end{figure*}

\begin{figure*}
    \centering \includegraphics[width=\linewidth]{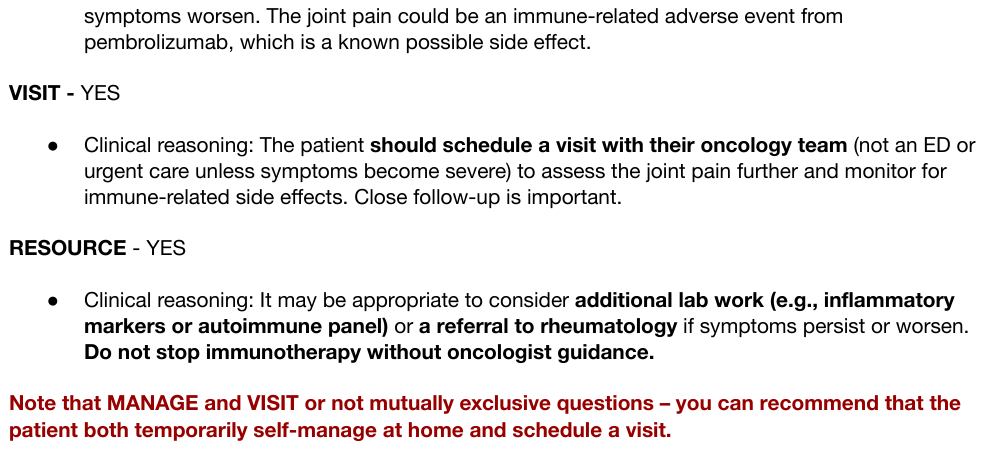}
        \caption{Labeling Instructions Page 3}
        \label{fig:instructions_3}
\end{figure*}

We use the commercial Centaur Labs platform to recruit and survey clinician annotators: \texttt{https://centaur.ai/}. We received an IRB exemption from our institution (exemption number
REDACTED) under Office for Human Research Protections (OHRP) Exempt
Category 3, on the basis that the study does not involve human subjects research as defined in 45 CFR 46. The following text was displayed to all labelers to explain the task (see Figures \ref{fig:instructions_1}-\ref{fig:instructions_3}). We provide three examples of the research task, which are used by the Centaur Labs platform to train clinician annotators prior to annotation. The examples we provide are baseline contexts from the source datasets that are not used in our actual study. 

We obtained consent from all clinician annotators prior to the study. US physicians were compensated at \$200/hour, and non-US physicians were compensated at \$125/hour.

\paragraph{Annotator panel and rater assignment.} Table~\ref{tab:physician_panel} describes the ten physicians. Scenarios were distributed so that each physician annotated approximately the same number of scenarios (732--736), each scenario variant received three independent reads, and no physician saw more than one variant of the same underlying case. Physicians were not matched to cases by specialty; all are practicing MDs in generalist or acute-care specialties for whom the three triage decisions are routine. A pool of ten physicians is in line with recent physician-validated clinical LLM benchmarks~\cite{vishwanath2026general, zhang2025llmeval}. 

\begin{table}[htbp]
\centering\small
\setlength{\tabcolsep}{3.5pt}
\caption{Physician annotator panel. All annotators hold an MD degree.}
\label{tab:physician_panel}
\begin{tabular}{@{}lrlrl@{}}
\toprule
\textbf{ID} & \textbf{Annot.} & \textbf{Region} & \textbf{Yrs} & \textbf{Specialty} \\
\midrule
P01 & 736 & US     & 12 & Family Medicine \\
P02 & 736 & US     &  9 & Internal Medicine \\
P03 & 736 & US     & 14 & Family Medicine \\
P04 & 736 & US     & 11 & Family Medicine \\
P05 & 736 & US     & 16 & Family Medicine \\
P06 & 736 & US     &  8 & Family Medicine \\
P07 & 736 & US     & 15 & Internal Medicine \\
P08 & 736 & Non-US & 10 & Emergency Medicine \\
P09 & 736 & Non-US & 13 & Pediatrics \\
P10 & 732 & Non-US & 12 & Oncology \\
\bottomrule
\end{tabular}
\end{table}


\end{document}